\documentclass[letterpaper, 10 pt, conference]{ieeeconf}  

\IEEEoverridecommandlockouts                              
\usepackage{mathtools,amssymb,commath}
\usepackage{amsmath}
\usepackage{graphicx,import}
\usepackage{verbatim}
\usepackage{color}
\usepackage{hyperref}
\usepackage{microtype}
\hypersetup{colorlinks=true}   

\newcommand{\nJoints}{n}
\newcommand{\nJets}{{n_p}}

\DeclareMathOperator{\skewMatrix}{S}
\DeclareMathOperator*{\argmin}{argmin}

\newcommand{\nR}[1]{\mathbb{R}^{#1}} 

\newcommand{\jetForces}{{F}}
\newcommand{\jetInt}{{T}}
\newcommand{\jetIntDot}{\dot{\jetInt}}
\newcommand{\jetAxis}{{a}}
\newcommand{\jetArm}{{r}}
\newcommand{\jetMapping}{{A}}
\newcommand{\rpm}{{\text{RPM}}}
\newcommand{\rpmThr}{{\rpm_{\text{thr}}}}

\newcommand{\baseFrame}{\mathcal{B}}
\newcommand{\comFrame}{\mathcal{G}}
\newcommand{\inertialFrame}{\mathcal{I}}

\newcommand{\centrFrame}{\comFrame[\inertialFrame]}
\newcommand{\centrBodyFrame}{\comFrame[\baseFrame]}

\newcommand{\jacobian}{{J}}

\newcommand{\CMM}{{J}_\comFrame}

\newcommand{\frameLinVel}{{{{v}}}}

\newcommand{\jointTorques}{{{\tau}}}
\newcommand{\massMatrix}{{M}}
\newcommand{\coriolis}{{C}}
\newcommand{\gravityForces}{{G}}

\newcommand{\totalInertia}{{I}}
\newcommand{\gravity}{g}
\newcommand{\gravForce}{F_G}
\newcommand{\totalMass}{m}

\newcommand{\jointPos}{{{s}}}
\newcommand{\jointVel}{\dot{{\jointPos}}}

\newcommand{\basePos}{{p}_{\baseFrame}}
\newcommand{\baseRot}{{R}_{\baseFrame}}

\newcommand{\baseLinVel}{{v}_{\baseFrame}}

\newcommand{\baseAngVel}{{\omega}_{\baseFrame}}

\newcommand{\systemPos}{{q}}
\newcommand{\systemVel}{{\nu}}
\newcommand{\systemAcc}{\dot{{\nu}}}

\newcommand{\momentum}{{L}}
\newcommand{\momentumDot}{\dot{{\momentum}}}
\newcommand{\momentumDdot}{\ddot{{\momentum}}}
\newcommand{\momentumDdotMixed}{\ddot{{\momentum}}}
\newcommand{\linMom}{{l}}
\newcommand{\linMomDot}{\dot{{\linMom}}}
\newcommand{\linMomErr}{{\tilde{\linMom}}}
\newcommand{\linMomErrDot}{{\dot{\linMomErr}}}
\newcommand{\linMomDdot}{\ddot{{\linMom}}}
\newcommand{\angMom}{\text{w}}
\newcommand{\angMomDot}{\dot{{\angMom}}}
\newcommand{\angMomDdot}{\ddot{{\angMom}}}

\newcommand{\manipMatrix}{{\Lambda}}
\newcommand{\constrSelfColl}{{h}_{\text{self-coll}}}
\newcommand{\constrJetColl}{{h}_{\text{jet-coll}}}
\newcommand{\centerSphere}{{c}}
\newcommand{\centerCone}{{p}}
\newcommand{\nullCone}{{N}_\jetAxis}
\newcommand{\ControlMatrix}{{B}}
\newcommand{\ControlBias}{{b}}
\newcommand{\inputControlOpti}{{x}}
\newcommand{\inputControl}{{u}}
\newcommand{\inputControlInt}{{I_{\inputControl}}}
\newcommand{\upperBoundU}{\inputControl^{max}}  
\newcommand{\lowerBoundU}{\inputControl^{min}}  
\newcommand{\upperBoundUInt}{\inputControlInt^{max}}  
\newcommand{\lowerBoundUInt}{\inputControlInt^{min}}  
\newcommand{\upperBoundUOpti}{\inputControlOpti^{max}}  
\newcommand{\lowerBoundUOpti}{\inputControlOpti^{min}}  
\newcommand{\WeightLinMom}{{W}_\linMom}
\newcommand{\WeightAngMom}{{W}_\angMom}
\newcommand{\WeightOptiError}{{W}_\inputControlOpti}
\newcommand{\WeightOptiManip}{{W}_\manipMatrix}
\newcommand{\WeightLinMomBar}{\bar{W}_\linMom}
\newcommand{\WeightAngMomBar}{\bar{W}_\angMom}
\newcommand{\WeightPost}{{W}_\jointPos}

\newcommand{\KiMomentum}{{K}_I}
\newcommand{\KpMomentum}{{K}_P}
\newcommand{\KdMomentum}{{K}_D}
\newcommand{\KpJoints}{{K}^s_{P}}

\usepackage{mathtools,amsthm}
\usepackage{algorithm, algpseudocode}

\title{\LARGE \bf Failure Detection and Fault Tolerant Control \\ of a Jet-Powered Flying Humanoid Robot
}

\author{Gabriele Nava $^{1}$, Daniele Pucci $^{1,2}$
\thanks{$^{1}$Artificial and Mechanical Intelligence Laboratory, Fondazione Istituto Italiano di Tecnologia, via San Quirico 19D, Genoa, Italy. {Email addresses: \tt\small firstname.surname@iit.it}}
\thanks{$^{2}$School of Computer Science, University of Manchester,
        Manchester M13 9PL, United Kingdom}
}

\begin{document}

\maketitle
\thispagestyle{empty}
\pagestyle{empty}

\begin{abstract}
Failure detection and fault tolerant control are fundamental safety features of any aerial vehicle. 
With the emergence of complex, multi-body flying systems such as jet-powered humanoid robots, it becomes of crucial importance to design fault detection and control strategies for these systems, too. In this paper we propose a fault detection and control framework for the flying humanoid robot iRonCub in case of loss of one turbine. The framework is composed of a failure detector based on turbines rotational speed, a momentum-based flight control for fault response, and an offline reference generator that produces far-from-singularities configurations and accounts for self and jet exhausts collision avoidance. Simulation results with Gazebo and MATLAB prove the effectiveness of the proposed control strategy.
\end{abstract}

\section{INTRODUCTION}
\label{sec:intro}


Promising research projects are focusing on the design of complex, multi-body flying robots mixing aerial, terrestrial and manipulation capabilities. In this context, the application of state-of-art Fault Tolerant Control (FTC) strategies in response to actuation failures is impaired by the complexity of these systems' internal dynamics.
In this paper we present a strategy to address fault detection and control for a multi-body aerial system, and demonstrate the effectiveness of our framework with realistic flight simulations. 
 
Fault control strategies for aerial systems can be roughly divided into two categories: \emph{passive} and \emph{active} control design \cite{Yin2016},\cite{Zhang2003},\cite{fourlas2021}. Passive strategies encompass all cases where the fault is managed without reconfiguration of the control algorithm. For example, airplanes and spacecraft can handle failures passively by duplicating sensors and actuators; this may not be possible on small aerial vehicles because of weight and space limitations. The controller of the latter is then designed to be robust to a certain number of expected failures. State-of-art passive fault control strategies for small aerial systems include: PID parameters optimization, sliding mode control, neural networks \cite{jun2010},\cite{gong2010},\cite{mallavalli2019},\cite{yu2021}. The main drawback of these passive control designs it that they guarantee robustness only for predefined set of failures. Also, achieving robustness to certain failures is possible only at the cost of decreased nominal performances \cite{Yin2016}. 
\begin{figure}[ht]
  \centering
  \includegraphics[width=1\columnwidth]{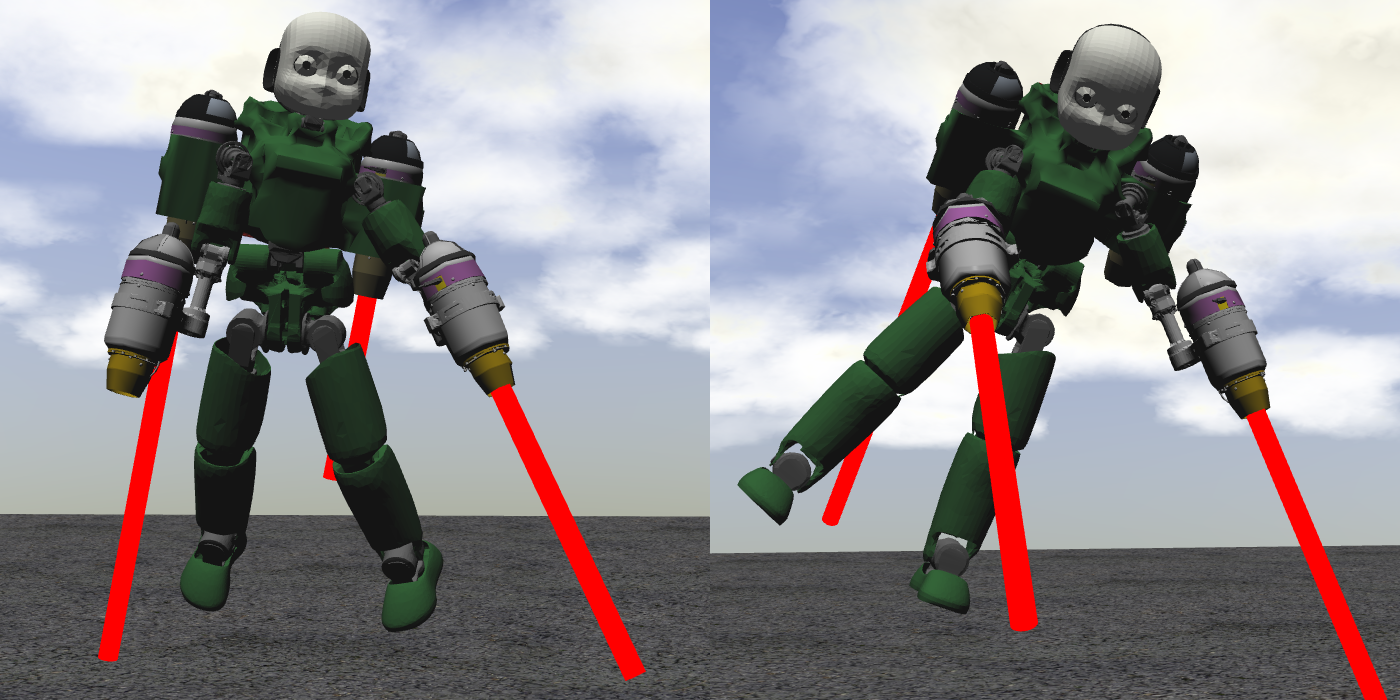}
  \vspace{-0.2cm}
  \caption{The iRonCub robot hovering in simulation with one faulty turbine. Left: right arm turbine fault; Right: left back turbine fault.}
  \label{fig:faultSimulation}
  \vspace{-0.4cm}
\end{figure} 

On the other hand, active fault control strategies reconfigure the controller in case of fault, to guarantee stability and acceptable performances in the new operating conditions. Active FTC requires the usage of a \emph{fault detection} system to properly detect the occurred fault and trigger control reconfiguration \cite{Yin2016}. Examples of active fault controllers include gain scheduling PID, sliding mode and model predictive control, adaptive control, H-infinity control \cite{Sadeghzadeh2013}, \cite{Cheng2018}, \cite{Yu2015}, \cite{zhong2018}. \cite{Nguyen2018}. 

Most of the above-mentioned active and passive control strategies are developed for small, single body aerial systems such as quadrotors, fixed-wing, and hexacopters \cite{fourlas2021}. Novel research directions, however, are trying to design UAVs with several degrees of freedom, which is a fundamental step for providing flying robots with manipulation or terrestrial locomotion capabilities. These flying multibody robots include aerial manipulators, hexapod-quadrotor designs, hybrid terrestrial and aerial quadrotors, and humanoid robots with thrusters to enhance bipedal locomotion \cite{kim2021}, \cite{huang2017}, \cite{ruggiero2018}, \cite{kalantari2013}, \cite{pitonyak2017}. Effective fault control design of such multibody systems is non-trivial, because of the inherent complexity of their internal dynamics, increased weight and/or relatively slow actuators dynamics.  These challenges call for the design of new, dedicated fault detection and control strategies for complex multi-body systems. Only few state-of-art work started to address the problem, focusing mainly on fault control design for aerial manipulators \cite{pose2022}, \cite{garimella2021}. 

In this paper we propose a fault detection and control framework for the jet-powered humanoid robot iRonCub \cite{pucci2017fly}, \cite{mohamed2020}, \cite{Bartolozzi2017}. As highlighted in Fig. \ref{fig:faultSimulation}, four jet engines are mounted on the robot arms and shoulders, and used to lift the robot from the ground. We focus on the case in which the failure is due to a malfunctioning of one thruster.
The proposed approach includes: 1) a fault detection system that allows to detect if a thruster sustained a failure; 2) modifications of the existing iRonCub momentum-based flight controller to become aware of the occurred fault, and reconfigure the robot to retain stable flight; 3) an offline planner that generates updated references for the robot's attitude and internal configuration in case of faulty actuators. The planner provides far from singularities references while avoiding self and jet exhausts collisions. The framework is validated with simulations in MATLAB and Gazebo.
  
The reminder of the paper is organized as follows: Sec. \ref{sec:backgorund} recalls the system modeling and the momentum-based control developed in our previous work \cite{pucci2017fly}, \cite{nava2018}; Sec. \ref{sec:problem_statement} describes the limits of the flight controller in case of turbine failure; Sec. \ref{sec:fault_detection} details the fault identification framework, while Sec. \ref{sec:fault_control} presents our fault control design; Sec. \ref{sec:ref_gen} describes the offline reference generator. Results are presented in Sec. \ref{sec:simulations}. Final remarks and possible improvements conclude the paper.

\section{BACKGROUND}
\label{sec:backgorund}
  
\subsection{System Modeling}  
  
The flying system under study is composed of $\nJoints + 1$ bodies (links) connected by $\nJoints$ joints with 1 DoF each, and by $\nJets$ thrusters rigidly mounted on the robot links. The system's kinematics and dynamics is modeled by resorting to the \emph{floating base} formalism \cite{Featherstone2007}. More specifically, we define the robot configuration as $\systemPos := (\basePos, \baseRot, \jointPos) \in \nR{3} \times SO(3) \times \nR{\nJoints}$, which is composed by the position $\basePos \in \nR{3}$ and orientation $\baseRot \in SO(3)$ of one of the robot's links, denoted as \emph{base link} $\baseFrame$, w.r.t. the inertial frame $\inertialFrame$, and the joint positions $\jointPos \in \nR\nJoints$ characterizing the \emph{internal shape} of the robot. The base rotation is represented with rotation matrices to avoid singularities that may arise when choosing a minimal representation of rotations with Euler angles. The system velocities are defined by the \emph{algebra} of $\systemPos$, and given by: $\systemVel := (\baseLinVel, \baseAngVel, \jointVel) \in \nR{6+\nJoints}$, where $\baseLinVel$, $\baseAngVel \in \nR{3}$ are the base link linear and angular velocities w.r.t. the inertial frame, and $\jointVel \in \nR\nJoints$ are the joint velocities. 

We write the system's equations of motion via  Euler-Poincar\'e formalism \cite{Marsden2010}:
\begin{equation}
\label{eq:sys_dynamics}
  \massMatrix(\systemPos)\systemAcc + \coriolis(\systemPos,\systemVel)\systemVel+\gravityForces(\systemPos) = \begin{bmatrix} 0_6 \\ \jointTorques \end{bmatrix} +\sum_{k=1}^{\nJets}\jacobian_k^{\top} \jetForces_k,
\end{equation}
where $\massMatrix$, $\coriolis \in \nR{\nJoints+6\times \nJoints+6}$ represent the inertia and Coriolis matrices, $\gravityForces \in \nR{\nJoints+6}$ is the gravity vector, $\jointTorques \in \nR{\nJoints}$ are the internal actuation torques and $\jetForces_k \in \nR{3}$ is the $k^{th}$ external force acting on the system. We assume that the forces acting on the robot during flight are only the thrust forces generated by the thrusters. This assumption is reasonable in case of low flight velocity and absence of wind, otherwise aerodynamics forces must also be considered. Each thrust force can be represented as $\jetForces_k={}^\inertialFrame \jetAxis_k(\systemPos) \jetInt_k$, where ${}^\mathcal{I}\jetAxis_k \in \nR{3}$ is the axis of the force, while $T_k \in \nR{}$ is the thrust intensity. The jacobian $\jacobian_k(\systemPos) \in \nR{3 \times \nJoints}$ is the map between system velocities $\systemVel$ and the linear velocity $\frameLinVel_k$ of the $k^{th}$ thrust application point. 

\subsection{Momentum-based Control for Flight}
\label{subsec:control_background}

Writing efficient control algorithms to stabilize the full system dynamics Eq. \eqref{eq:sys_dynamics} is, in general, not an easy task. Depending on the number of thrusters and how their axes are oriented, the system may be \emph{underactuated}, and this complexifies the control design \cite{Acosta05}. 

A common design choice is to select a control output of a dimension smaller than the full system dynamics, but representative of the \emph{overall} system stability during flight. In this sense, recall that the rate of change of \emph{centroidal momentum} equals the sum of all external forces and moments acting on the system, i.e.:
\begin{equation}
\label{eq:centr_mom_dot}
  {}^{\centrFrame}\momentumDot = \begin{bmatrix} {}^{\centrFrame}\linMomDot \\ {}^{\centrFrame}\angMomDot \end{bmatrix} = \begin{bmatrix} \sum_{k = 1}^{\nJets} \jetForces_k + \totalMass\gravity e_3 \\ \sum_{k = 1}^{\nJets} \skewMatrix(\jetArm_k)\jetForces_k
  \end{bmatrix},
\end{equation}
with ${}^{\centrFrame}\momentum \in \nR{6}$ the centroidal momentum, that is, the momentum of the system expressed w.r.t. a reference frame with origin at the system's center of mass and orientation of the inertial frame. We indicate with ${}^{\centrFrame}\linMom$, ${}^{\centrFrame}\angMom \in \nR{3}$ the linear and angular momentum of the system in centroidal coordinates, respectively. $\skewMatrix(\cdot)$ is the skew operator in $\nR{3}$ and $\jetArm_k$ is the distance between system's center of mass and the $k^{th}$ thurst application point. $\totalMass\gravity e_3 \in \nR{3}$ is the gravity force, with $\totalMass \in \nR{}$ the total mass of the robot. 

Only the magnitude of the 3D thrust forces $\jetForces_k$ can be directly  actuated by the thrusters: to orient the thrust force axes one must resort to the internal kinematics of the robot. To better highlight the (nonlinear) dependency of the thrust axes from the joint positions, we recall that $\jetForces_k={}^\inertialFrame \jetAxis_k(\systemPos) \jetInt_k$ and rewrite Eq. \eqref{eq:centr_mom_dot} as:
\begin{IEEEeqnarray}{LCL}
\label{eq:centr_mom_dot_simpl}
  {}^{\centrFrame}\momentumDot & = & \begin{bmatrix} \jetAxis_1 & ... & \jetAxis_\nJets \\ \skewMatrix(\jetArm_1)\jetAxis_1 & ... & \skewMatrix(\jetArm_\nJets)\jetAxis_\nJets \end{bmatrix} \begin{bmatrix} \jetInt_1 \\ ... \\ \jetInt_\nJets \end{bmatrix} + \begin{bmatrix} \totalMass\gravity e_3 \\ 0_{3,1} \end{bmatrix} \\
 & = & \jetMapping (\systemPos) \jetInt + \gravForce, \nonumber
\end{IEEEeqnarray}
where $\jetMapping \in \nR{6 \times \nJets}$ and $\gravForce \in \nR{6}$. $\jetInt \in \nR{\nJets}$ stacks all thrusts magnitude in a single vector.


On jet-powered robots, the relatively slow turbines dynamics often impairs the possibility to stabilize the momentum rate of change Eq. \eqref{eq:centr_mom_dot_simpl} relying only on thrust intensities $\jetInt$. To overcome this problem, joint positions could be used as additional control input to modify the thrust axes $\jetAxis(\systemPos)$, but they appear nonlinearly in Eq. \eqref{eq:centr_mom_dot_simpl}.
To handle the nonlinearity we decided to apply \emph{relative degree augmentation} and choose as control output the \emph{momentum acceleration}:
\begin{IEEEeqnarray}{LCL}
\label{eq:centr_mom_ddot}
  \momentumDdotMixed  := \begin{bmatrix}
                          {}^{\centrFrame}\linMomDdot \\
                          {}^{\centrBodyFrame}\angMomDdot
                         \end{bmatrix}  = \begin{bmatrix}
                                          \frac{\partial{({}^{\centrFrame}\linMomDot})}{\partial{\jetInt}}\jetIntDot + \frac{\partial{({}^{\centrFrame}\linMomDot})}{\partial{\systemPos}}\systemVel\\
                                          \frac{\partial{({}^{\centrBodyFrame}\angMomDot})}{\partial{\jetInt}}\jetIntDot + \frac{\partial{({}^{\centrBodyFrame}\angMomDot})}{\partial{\systemPos}}\systemVel
                                          \end{bmatrix}.
\end{IEEEeqnarray}
We also express the angular momentum in \emph{body} coordinates, ${}^{\centrBodyFrame}\angMom$, that is, the angular momentum w.r.t. a frame attached to the robot center of mass, with the orientation of the base frame. The angular momentum can be easily mapped from centroidal to body coordinates using the base rotation ${}^{\centrBodyFrame}\angMom = \baseRot^\top {}^{\centrFrame}\angMom$. In this coordinates representation, the total inertia of the robot $\totalInertia \in \nR{3 \times 3}$ only depends on the internal robot shape, i.e. $\totalInertia = \totalInertia(\jointPos)$, and this property facilitates the design of the attitude control.  We select as control input of Eq. \eqref{eq:centr_mom_ddot} the rate of change of thrust magnitude and the joint velocities, i.e. $\inputControl := (\jetIntDot, \jointVel)$, as they both appear linearly in the momentum acceleration equations. The desired closed loop dynamics for control output \eqref{eq:centr_mom_ddot} is selected as:
\begin{IEEEeqnarray}{LCL}
\label{eq:centr_mom_ddot_star}
   \momentumDdot^*  = \begin{bmatrix} {}^{\centrFrame}\linMomDdot^* := {}^{\centrFrame}\linMomDdot^d - \KdMomentum                                    \linMomErrDot -\KpMomentum \linMomErr - \KiMomentum \int_0^t \linMomErr \\
                                       {}^{\centrBodyFrame}\angMomDdot^*             
                        \end{bmatrix} 
\end{IEEEeqnarray}
where $\KdMomentum, \KpMomentum, \KiMomentum \in \nR{3 \times 3}$ are positive definite gains matrices, $\linMomErr = {}^{\centrFrame}\linMom - {}^{\centrFrame}\linMom^d$, and ${}^{\centrFrame}\linMom^d$ the reference linear momentum provided by, e.g., a planner or user inputs. The desired angular momentum acceleration ${}^{\centrBodyFrame}\angMomDdot^*(\systemPos, \jetInt, \baseAngVel)$ is the output of a dedicated attitude controller described in previous work \cite{nava2018}.

To find the optimal control input $\inputControl^* := (\jetIntDot^*, \jointVel^*)$ we resort to Quadratic Programming (QP) optimization, and write the following minimization problem:
\begin{IEEEeqnarray}{LCL}
\label{eq:qp_flight_basic}
  \inputControl^* & = & \argmin_{\inputControl := (\jetIntDot, \jointVel)} \frac{|\linMomDdot -\linMomDdot^*|_{\WeightLinMom}^2 + |\angMomDdot - \angMomDdot^*|_{\WeightAngMom}^2 + |\jointVel-\jointVel^*|_{\WeightPost}^2}{2} \\ 
  & s.t. & \nonumber \\
  & & \lowerBoundU \leq \inputControl \leq \upperBoundU \nonumber
\end{IEEEeqnarray}
where the terms in the cost function, weighted by the positive weights $\WeightLinMom, \WeightAngMom, \WeightPost$, are the linear and angular momentum tracking tasks (we omitted the coordinates frame representation ${}^{\centrFrame}$ and ${}^{\centrBodyFrame}$ for space limitations), and a postural task with $\jointVel^* = -\KpJoints (\jointPos - \jointPos^d)$ to resolve eventual actuation redundancy. $\lowerBoundU$ and $\upperBoundU$ are the inputs lower and upper bounds. For the linear momentum part, recall that the momentum can be written as a function of the system velocities via the \emph{centroidal momentum matrix} $\CMM$, i.e $\momentum = \CMM (\systemPos) \systemVel$ \cite{Orin2008}. We therefore need to handle the dependency on joint velocities in the term $\linMomErr$ of Eq. \eqref{eq:centr_mom_ddot_star} to avoid algebraic loops. For this reason, in a previous work we reformulated the linear momentum tracking problem from $\linMomDdot = \linMomDdot^*$ to $\ControlMatrix \inputControl + \ControlBias = 0$, with $\ControlMatrix(\systemPos, \jetInt) \in \nR{3 \times {\nJets+\nJoints}}$ and $\ControlBias(\systemPos, \jetInt, \baseAngVel, \baseLinVel) \in \nR{3}$, obtained by rewriting the original problem with the help of an auxiliary state variable \cite{pucci2017fly}. The new formulation also adds constraints to the gains matrices in Eq. \eqref{eq:centr_mom_ddot_star} that facilitate gain tuning. The optimal thrust rate of change and joint velocities $u^* = (\jetIntDot^*, \jointVel^*)$ obtained from the QP problem \eqref{eq:qp_flight_basic} are then achieved by means of a lower-level, faster thrust and torque controller. The baseline controller Eq. \eqref{eq:qp_flight_basic}  will be modified in Sec. \ref{sec:fault_control} to address fault control. More details on the flight control design, including its stability analysis, can be found in the already mentioned previous work \cite{nava2018},\cite{pucci2017fly}.

\section{PROBLEM STATEMENT}
\label{sec:problem_statement}

\subsection{Definition of Failure}

In this paper we assume that the failure is caused by a malfunction of the thrusters. In a jet-powered robot as the iRonCub, a thruster failure can be triggered by various reasons: presence of bubbles in the fuel line, reduced air flow at the turbine's inlet (which can lead to over-temperature issues), presence of dust or other particles that can damage the blades, and so on. We chose to focus on this type of failure as there is hope to retain flight stability once the fault is properly detected and isolated. Other failures, such as the loss of control of the robot joints or a shut down of the on-board laptop, would most likely lead to a catastrophic crash with no possibility of recovery. We concentrate on the case of \emph{complete fault}: the thruster suddenly shuts down, and it is not capable anymore to provide thrust force. Addressing a \emph{partial fault} is also possible, but this would require to design a more sophisticated anomaly detection algorithm which will be subject of future work.

\subsection{Limits of Momentum-Based Control for Fault Response}
\label{subsec:controller_limits}

If the flight controller Eq. \eqref{eq:qp_flight_basic} remains unaware of the occurred failure, there will be no guarantees that the optimal input $\inputControl^*$ still stabilizes the robot momentum. It is therefore necessary to implement a \emph{fault detection} algorithm that monitors the engines status and provides this information to the controller. We implemented our own fault detection strategy which is detailed in Sec. \ref{sec:fault_detection}.

Awareness of the failure can be integrated in two ways in the QP problem:
\begin{enumerate}
    \item completely remove the faulty thruster from control by reshaping the terms $\frac{\delta (\cdot)}{\delta \jetInt}$ and $\jetIntDot$ in Eq. \eqref{eq:centr_mom_ddot}. This solution however increases the probability of generating discontinuous solutions for Eq. \eqref{eq:qp_flight_basic}, which may trigger unstable robot behavior;
    \item modify the actuation bounds in Eq. \eqref{eq:qp_flight_basic} to saturate the thrust magnitude of the faulty actuator. This modification may be implemented with a sharp but continuous function, therefore reducing the probability of big discontinuities in the QP solution. However, being the control input the thrust derivative $\jetIntDot$, it is not straightforward to apply such bounds to Eq. \eqref{eq:qp_flight_basic}.
\end{enumerate}
In what follows we chose this second route and exploited a parametrization of the input bounds, that allows to also include limits on thrust magnitude and joints position in Eq. \eqref{eq:qp_flight_basic}, designed in previous work \cite[Ch.14]{nava2020_thesis}, \cite{Rapetti2020}. When a fault is detected, input bounds are modified accordingly so that the QP can account for the occurred failure. The parametrization is detailed in Sec. \ref{subsec:bounds_param}.

Let us now assume that fault detection is implemented and the controller is aware of the faulty actuator: if there is still actuation redundancy, infinite sets of input $\inputControl$ can be found that achieve $\momentumDdot = \momentumDdot^*$. The (unique) solution to Eq. \eqref{eq:qp_flight_basic} is selected among the infinite possible depending on the controller gains and references, the actuation bounds and the tasks with lower priority, e.g. the postural task. In particular, controller references such as the desired joints configuration and robot attitude greatly influence the choice of the optimal solution $\inputControl^*$. When a fault occurs, these references should be updated to new ones that better fit with the failure situation, to avoid the QP converging to configurations that are, in practice, not very robust or even unfeasible because of singularities and self-collisions. This paper enforces and automatizes the references generation process by designing an offline optimizer to provide references for the robot posture and attitude in case of thruster fault.
More details on the reference generator algorithm are presented in Sec. \ref{sec:ref_gen}.

\section{FAILURE DETECTION}
\label{sec:fault_detection}

Fault detection is fundamental for the design of any active fault control strategy. In this paper, we focus on turbines fault. Turbines are often equipped with sensors capable of reading their rounds per minute (RPM), therefore we decided to implement a RPM-based fault detection strategy. More specifically, the measured RPM of each turbine are compared with the reference ones. If the absolute value of the RPM error overcomes a user-defined threshold $\rpmThr$ for a time longer than $t_{\rpmThr}$, a possible fault is detected. If the RPM descend below the idle value, the fault detection strategy detects that the turbine is off. 
%
%
%
The reference rounds per minute $\rpm^*(j)$ are obtained by integrating the thrust intensities rate of change $\jetIntDot^*$ calculated with Eq. \eqref{eq:qp_flight_basic}, and applying a static mapping between thrust and RPM \cite{Momin2022}. The measured RPM (from simulation) are quantized with steps of $100$ RPM for consistency with the real measurements.

\begin{figure}[t]
   \begin{minipage}[c]{8.5cm}
     \centering
     \includegraphics[width=1\columnwidth]{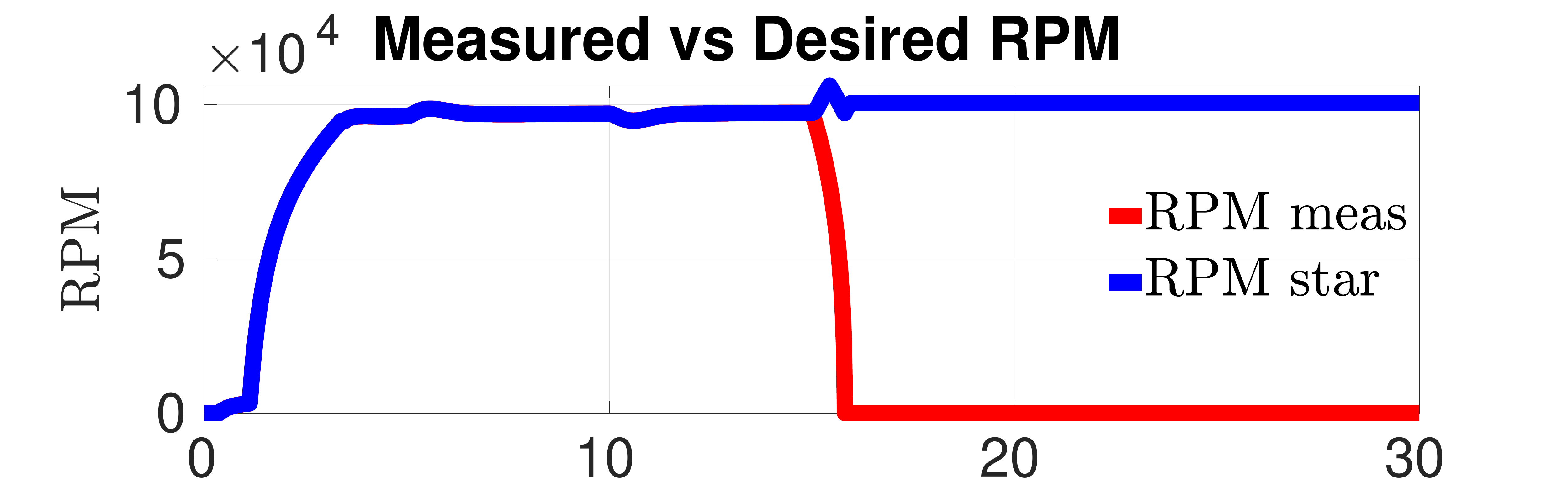}
     \vspace{-0.4cm}
   \end{minipage}
   \begin{minipage}[c]{8.5cm}
    \centering
    \includegraphics[width=1\columnwidth]{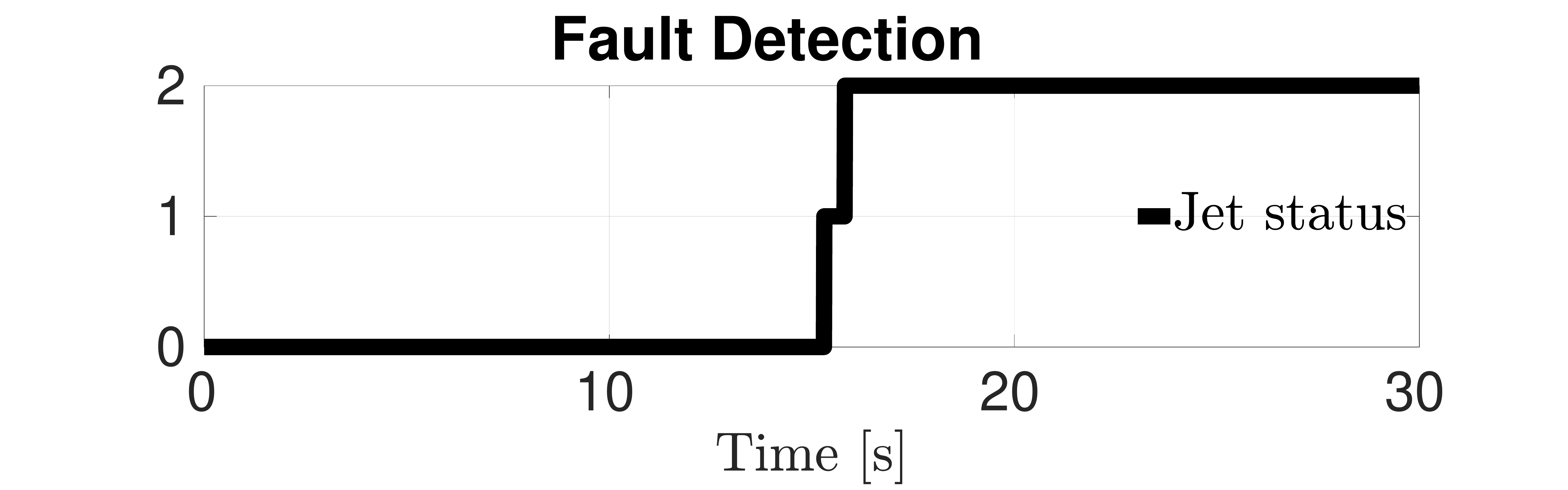}
    \vspace{-0.5cm}
    \caption{Performances of the fault detection strategy. Top figure: measured vs. desired RPM. Bottom figure: turbine status according to fault detection: 0 - no failure; 1 - a fault occurred; 2 - turbine is off.}
    \label{fig:faultDetection}
   \end{minipage}
   \vspace{-0.3cm}
  \end{figure}

Figure \ref{fig:faultDetection} shows the performances of the fault control strategy. Turbine fault occurs after $15$ $[s]$. It is detected in $0.3$ $[s]$ and the complete shutdown occurs $0.8$ $[s]$ after the fault started. Note that in simulation, idle RPM corresponds to zero RPM. The distinction between fault (state 1) and shutdown (state 2) may be useful to increase responsiveness of the controller to the failure. In what follows, we shall focus on the case of a complete shutdown and we postpone the analysis of partial failures to future work.

\section{FAULT CONTROL DESIGN}
  \label{sec:fault_control}
  
  Our fault control strategy relies on the momentum-based controller detailed in Sec. \ref{subsec:control_background}, with some modifications. First of all, we updated the input bounds $\lowerBoundU$ and $\upperBoundU$ to include also thrust and joint positions limits, and saturate the faulty turbine when a failure occurs. Secondly, we applied  weights scheduling to Eq. \eqref{eq:qp_flight_basic} to enforce controller robustness during the transient phase. 
  
  \subsection{Introduce Awareness of the Failure}
  \label{subsec:bounds_param}
  
  The idea is to include thrust and joint positions boundaries $\inputControlInt := (\jetInt, \jointPos)$ in the QP problem \eqref{eq:qp_flight_basic}, so that both limits on $\inputControlInt$ and $\inputControl$ are considered. We state our requirements as follows:
  \begin{itemize}
    \item define $\lowerBoundUInt$, $\upperBoundUInt$ as the boundaries of $\inputControlInt$; 
    \item at time instant $t$, if $\lowerBoundUInt \leq \inputControlInt(t) \leq \upperBoundUInt$, the current value of $\inputControlInt$ is respecting its limits. Therefore its derivative $\inputControl$ can be any value, provided that it respects the derivative bounds $\lowerBoundU \leq \inputControl(t) \leq \upperBoundU$;
    \item if $\inputControlInt(t) = \lowerBoundUInt$, the variable $\inputControlInt$ reached the lower bound. The derivative $\inputControl$ can be either $0$ or positive, but it cannot be negative (otherwise $\inputControlInt$ will further decrease, violating its limit). The bounds of $\inputControl$ need to be updated as $0 \leq \inputControl(t) \leq \upperBoundU$;
    \item the same approach can be used when $\inputControlInt(t) = \upperBoundUInt$ by setting $\lowerBoundU \leq \inputControl(t) \leq 0$;
  \end{itemize}
  Furthermore, assume that $\inputControlInt$ instantly overcomes the limits. This may occur on a real application because of unmodeled phenomena or disturbances that bring the state variables $\inputControlInt = (\jetInt, \jointPos)$ outside the limits. In this case, we must avoid $\inputControl$ to keep increasing (or decreasing), thus bringing $\inputControlInt$ far from the upper (lower) limit. Instead, $\inputControl$ must be strictly negative (positive) when $\inputControlInt$ overcomes the upper (lower) limit, in order to force $\inputControlInt$ to decrease (increase) and bringing it back in between the boundaries. As described in a previous work, all these requirements can be formulated analytically thanks to the properties of the hyperbolic tangent ($\tanh(\cdot)$, here abbreviated with $\text{th}(\cdot)$) \cite{Rapetti2020}:
  \begin{equation}
    \label{eq:input_bound_param}
    \text{th}(\epsilon_l(\inputControlInt-\lowerBoundUInt))\lowerBoundU \leq \inputControl \leq \text{th}(\epsilon_u(\upperBoundUInt-\inputControlInt))\upperBoundU,
  \end{equation}
  where $\epsilon_l$, $\epsilon_u$ are positive scalars defining the sharpness of the hyperbolic tangent. It can be verified that the new bounds \eqref{eq:input_bound_param} respect all the requirements listed above \cite[Ch. 14]{nava2020_thesis}, \cite{Rapetti2020}. We then substitute the input bounds of Eq. \eqref{eq:qp_flight_basic} with Eq. \eqref{eq:input_bound_param}.
  
  When a failure occurs, the part of the upper integral bound $\upperBoundUInt$ corresponding to the faulty turbine thrust is quickly driven to zero with a sharp but continuous profile. By consequence of Eq. \eqref{eq:input_bound_param}, the bounds on our QP input $\inputControl$ change, forcing the controller to stop relying on the faulty turbine for stabilization.
  
  \subsection{Weight Scheduling}
  \label{subsec:weights_scheduling}
  
  The transient phase between two flight configurations, i.e. from the robot hovering in normal conditions to stable flight with a faulty turbine, is critical for closed-loop system stability. During this transient, the instantaneous control action may lead the robot joints to hit the limits or get stuck in singularities that impair system convergence to a stable equilibrium. 
  To robustify our control design in this critical phase, we scale the QP weights of Eq. \eqref{eq:qp_flight_basic} $\WeightLinMom$ and $\WeightAngMom$ with a positive constant $\alpha > 0$: $\WeightLinMomBar = \frac{\WeightLinMom}{\alpha}$, $\WeightAngMomBar = \frac{\WeightAngMom}{\alpha}$. We linearly interpolate from  $\WeightLinMom$, $\WeightAngMom$ to $\WeightLinMomBar$, $\WeightAngMomBar$ and vice versa in the few seconds the failure occurs. 
  While tuning the parameter $\alpha$ to achieve the simulation results of Sec. \ref{sec:simulations}, we noted that allowing higher linear and angular momentum errors for a short time instant when failure is detected helps to avoid instabilities during transient phase. This result may be explicable as the momentum error when the failure occurs would be unavoidably higher than normal, and our (instantaneous) controller, normally fine-tuned to achieve momentum tracking with the highest accuracy possible, may react too aggressively to the error leading to instability. In all simulations we shall present in Sec. \ref{sec:simulations} the constant $\alpha$ is the same for both linear and angular momentum and also in case of arm or back turbine failure.
\section{OFFLINE REFERENCES GENERATOR}
  \label{sec:ref_gen}
  
  After recovering from a thruster failure, the robot converges to a configuration which guarantees stability of the momentum dynamics in the new operating conditions. If there is still actuation redundancy, it is useful to provide the system with updated references, to help the controller choose a more robust and performing solution and avoid possible instability issues as the ones hypothesized in Sec. \ref{subsec:controller_limits}. To this purpose, we designed an offline reference generator that computes optimized references for the robot attitude (represented by the base link orientation in roll-pitch-yaw, $\text{rpy}_\baseFrame$) and joints positions. More specifically, we solve the following nonlinear optimization problem:
  \begin{IEEEeqnarray}{LCL}
  \label{eq:offline_opti}
    \inputControlOpti^* & = & \argmin_{\inputControlOpti := (\text{rpy}_\baseFrame, \jointPos, \jetInt)} \frac{1}{2}{|\inputControlOpti -\inputControlOpti_0|_{\WeightOptiError}^2} + \frac{\WeightOptiManip}{\sqrt{\det{\manipMatrix^\top\manipMatrix}}}\IEEEyesnumber \\ 
    & s.t. & \nonumber \\
    & & |\momentumDot|^2 = 0 \nonumber \\
    & & \lowerBoundUOpti \leq \inputControlOpti \leq \upperBoundUOpti \nonumber \\
    & & \constrSelfColl(\inputControlOpti)  \geq 0  \nonumber \\
    & & \constrJetColl(\inputControlOpti)  \geq 0.  \nonumber
  \end{IEEEeqnarray}
  The cost function is composed of a first term penalizing solutions too different from the user-defined initial conditions $\inputControlOpti_0$,  and a second task that maximizes the determinant of the momentum \emph{manipulability ellipsoid}, here calculated as the matrix $\manipMatrix(\inputControlOpti) \in \nR{6 \times \nJoints + \nJets}$ that maps the control input of our QP controller Eq. \eqref{eq:qp_flight_basic}, namely $\inputControl := (\jetIntDot, \jointVel)$, into the momentum acceleration equations Eq. \eqref{eq:centr_mom_ddot}. Increasing manipulability should reduce the risk of reaching close to singular configurations that lead matrix $\manipMatrix$ to loose rank, impairing the controllability of the momentum acceleration through $\inputControl$. $\WeightOptiError$ and $\WeightOptiManip$ are positive weights.
  
\begin{figure}[t]
  \centering
  \includegraphics[width=1\columnwidth]{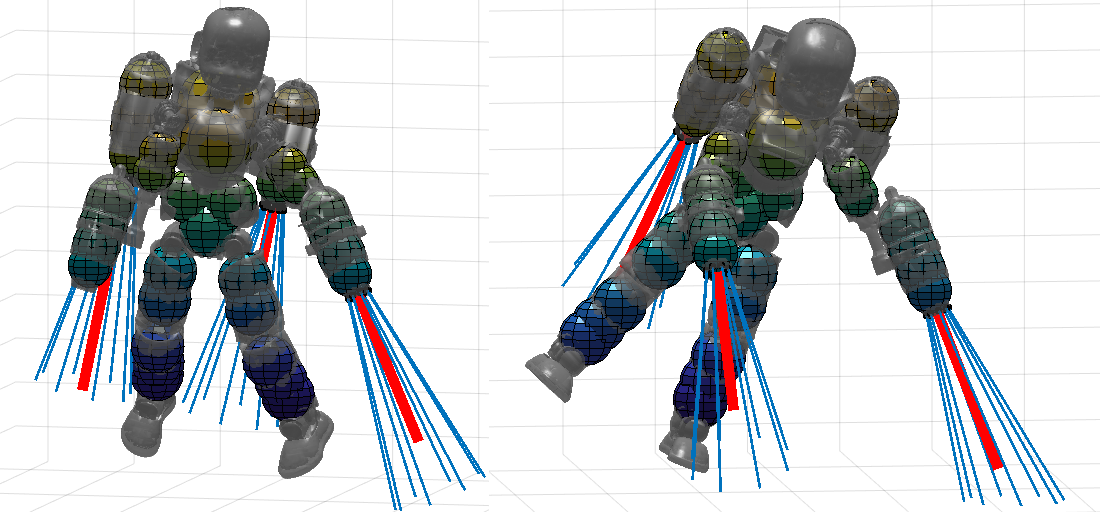}
  \vspace{-0.4cm}
  \caption{Optimized attitude, posture and thrust intensities in case of arm turbine failure (left) and back turbine failure (right). The colored spheres approximate robot links for self-collision avoidance. The blue lines parametrize the cones of jet exhausts.}
  \label{fig:faultOptimization}
  \vspace{-0.4cm}
\end{figure}

  Among the constraints, we require the square norm of momentum rate of change to be equal to zero to achieve static balance of forces and moments in the optimized hovering posture. We also include bounds on attitude, joint positions and thrust intensities via the variables $\lowerBoundUOpti$ and $\upperBoundUOpti$. The last two constraints represent 
  self and jet exhausts collision avoidance. Self collision is implemented by using multiple spheres to approximate each of the robot links (See also Fig. \ref{fig:faultOptimization}). We then check that the square distance among the centers of all pairs of spheres is greater than the sum of their radius, i.e. $ \constrSelfColl^{(i,j)}(\inputControlOpti) := |\centerSphere_i - \centerSphere_j |^2 - (\rho_i + \rho_j)^2 \geq 0$, with $\centerSphere_{i(j)}$ the center of the $i$-th ($j$-th) sphere in world coordinates, $\rho_{i(j)}$ the sphere radius, and $i$, $j$ the indices of two spheres that do not belong to the same link. Jet exhausts are approximated with lines shaping a cone that branches off the turbines nozzles with an inclination of $10~\text{[deg]}$, estimated empirically. The constraint ensures that the minimum distance between each line and the center of the link spheres is greater than the sphere radius:
  $ \constrJetColl^{(i,k)}(\inputControlOpti) := (\centerCone_k - \centerSphere_i)^\top \nullCone (\centerCone_k - \centerSphere_i) - \rho_i^2 \geq 0$, where $i$ are all spheres belonging to the robot lower body links, and $k$ all lines composing the cones. $\centerCone_k$ is a point that belongs to the $k$-th line, while $\nullCone = 1_3 - \jetAxis_k^\top\jetAxis_k$ projects the distance between sphere and line in the direction orthogonal to the line axis $\jetAxis_k$ (computing the minimum distance).
   
  The nonlinear optimization problem Eq. \eqref{eq:offline_opti} is solved in MATLAB using \emph{fmincon} with interior-point algorithm. The optimized references in case of right arm and left back turbine fault are depicted in Fig. \ref{fig:faultOptimization}. They have been provided to the flight controller and tested in simulation to improve performances and stability during flight.
\section{SIMULATION RESULTS}
\label{sec:simulations}

\subsection{Simulation Environment}

Our fault control is implemented in MATLAB-Simulink, and runs at a fixed-step frequency of $100~\text{[Hz]}$. The robot is simulated with Gazebo \cite{Koenig04}. The simulator offers different physic engines to integrate the system's dynamics. Among all the possibilities, we chose the Open Dynamics Engine (ODE), that uses a fixed step semi-implicit Euler integration scheme, with a simulation time step of $1$ [ms]. Gazebo and MATLAB communicate using YARP as middleware \cite{Metta2006}.

\subsection{Flight Scenario}

The simulated scenario consists in the robot taking off and hovering with four turbines, until one of the jets stops because of a failure (alternatively, the right arm or the left back turbine. Repeating the test with a failure on the remaining turbines is redundant because of the symmetry of joints and jets configuration). The robot then recovers from the failure, keeps hovering, and starts moving following a scripted trajectory. The trajectory includes up and down motion, rotation of the robot along its longitudinal axis and forward motion. The accompanying video shows more in detail how the trajectory looks like. Each simulation is performed $10$ times to verify its repeatability and robustness.

\begin{figure}[t]
   \begin{minipage}[c]{4.25cm}
     \centering
     \includegraphics[height=2.5cm, width=0.9\columnwidth]{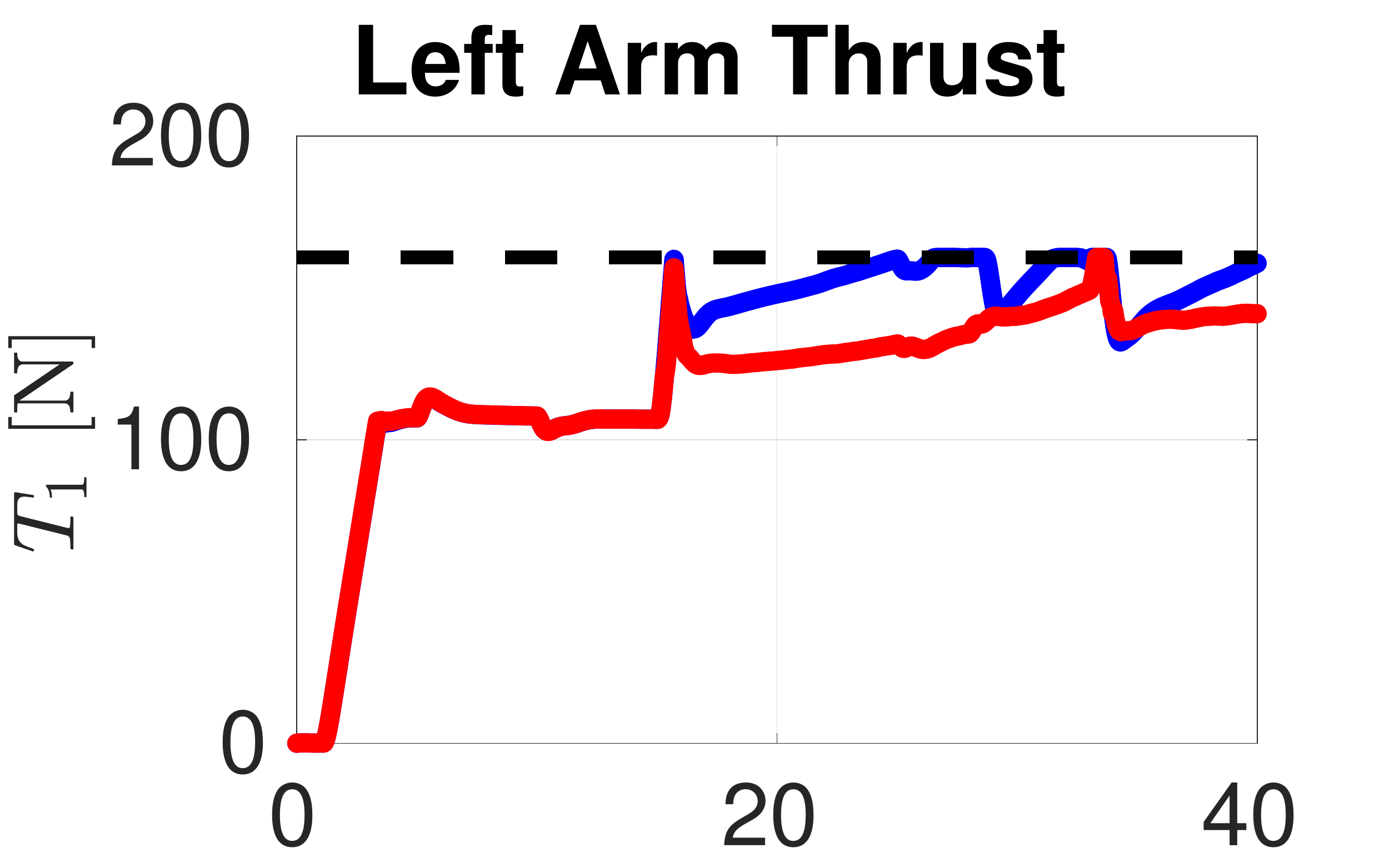}
     \hspace{-0.5cm}
   \end{minipage}
   \begin{minipage}[c]{4.25cm}
     \includegraphics[height=2.5cm, width=0.9\columnwidth]{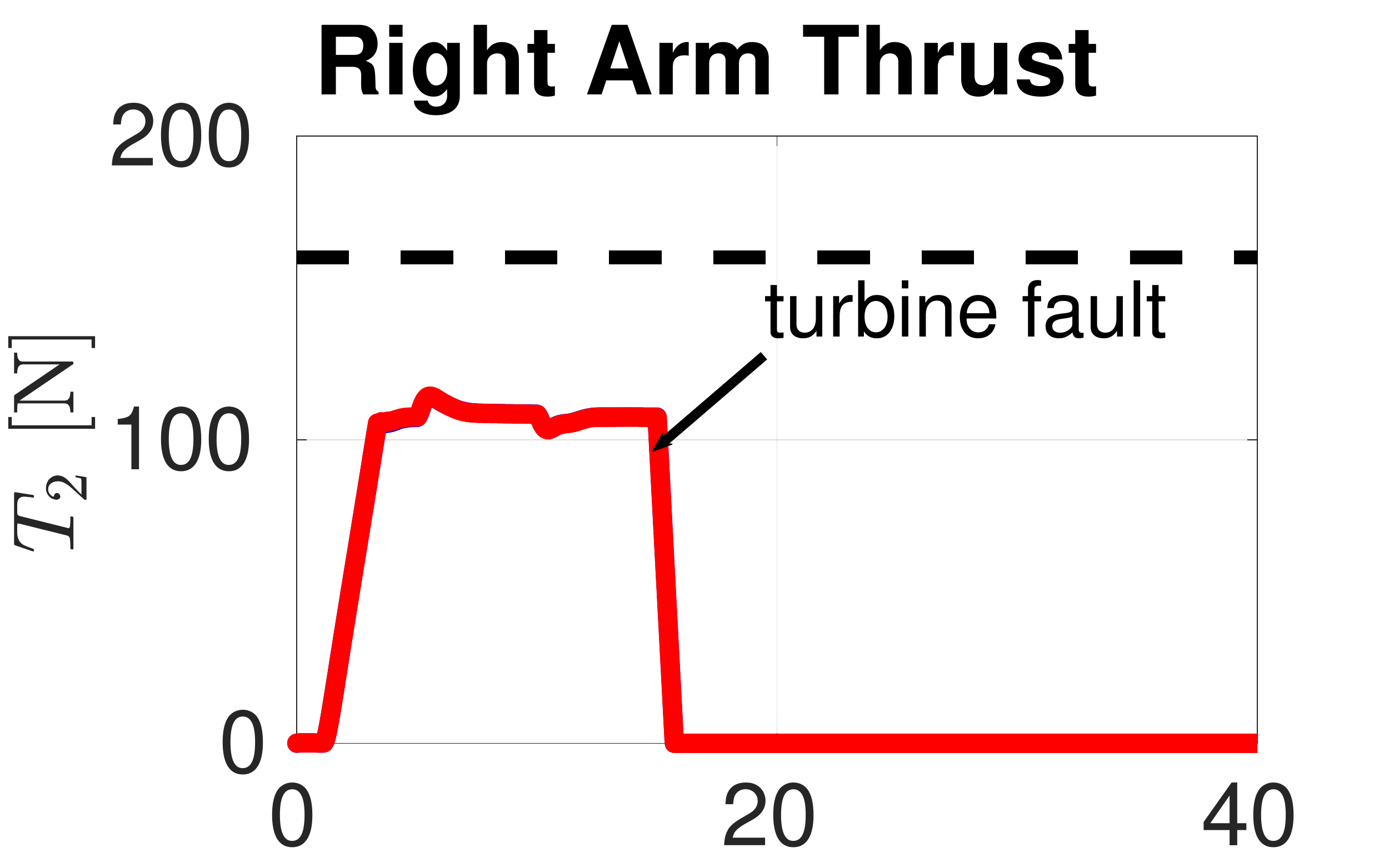}
   \end{minipage}
   \begin{minipage}[c]{4.25cm}
     \centering
     \includegraphics[height=2.95cm, width=0.9\columnwidth]{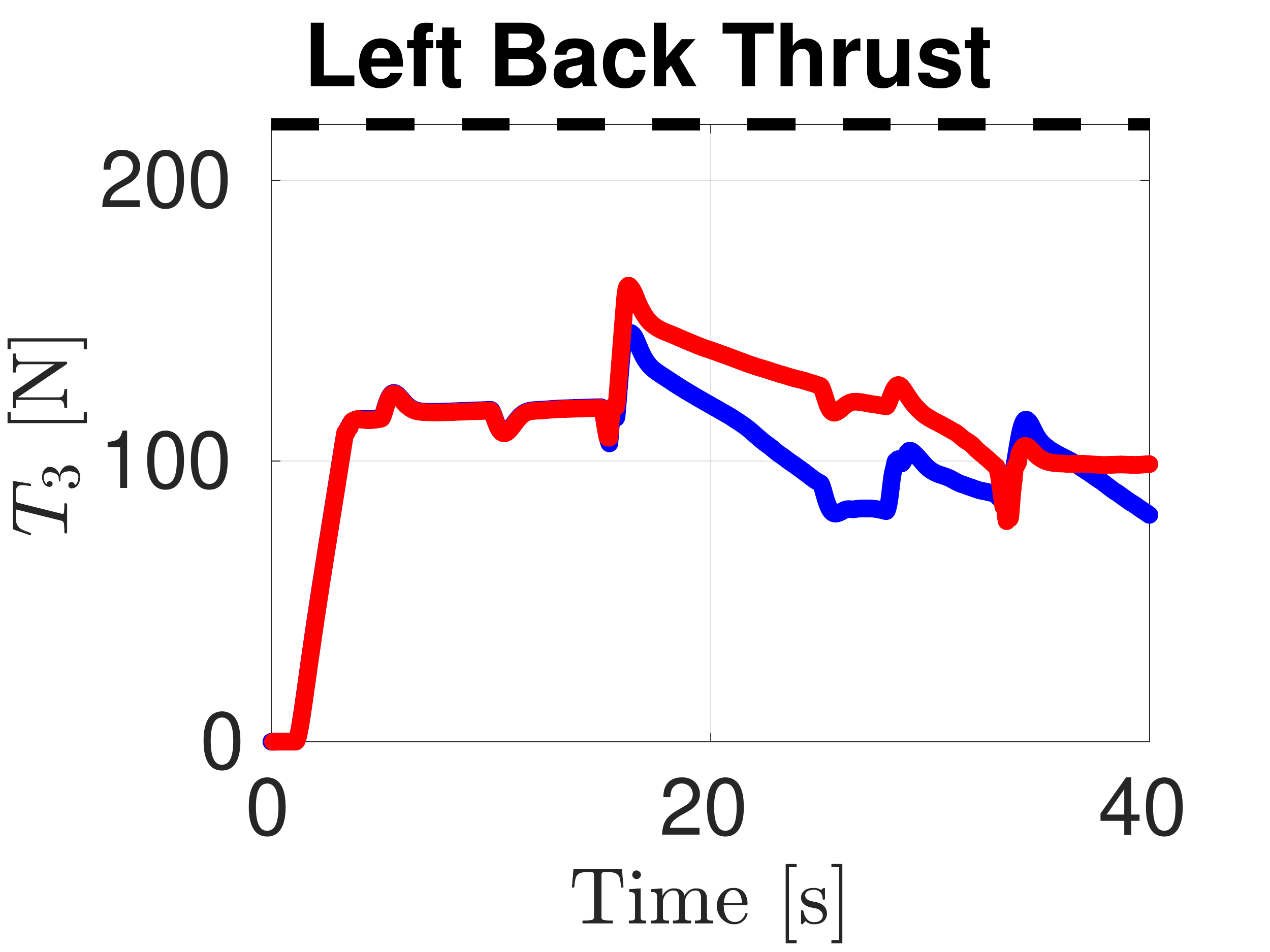}
     \hspace{-0.5cm}
   \end{minipage}
   \begin{minipage}[c]{4.25cm}
     \includegraphics[height=2.95cm, width=0.9\columnwidth]{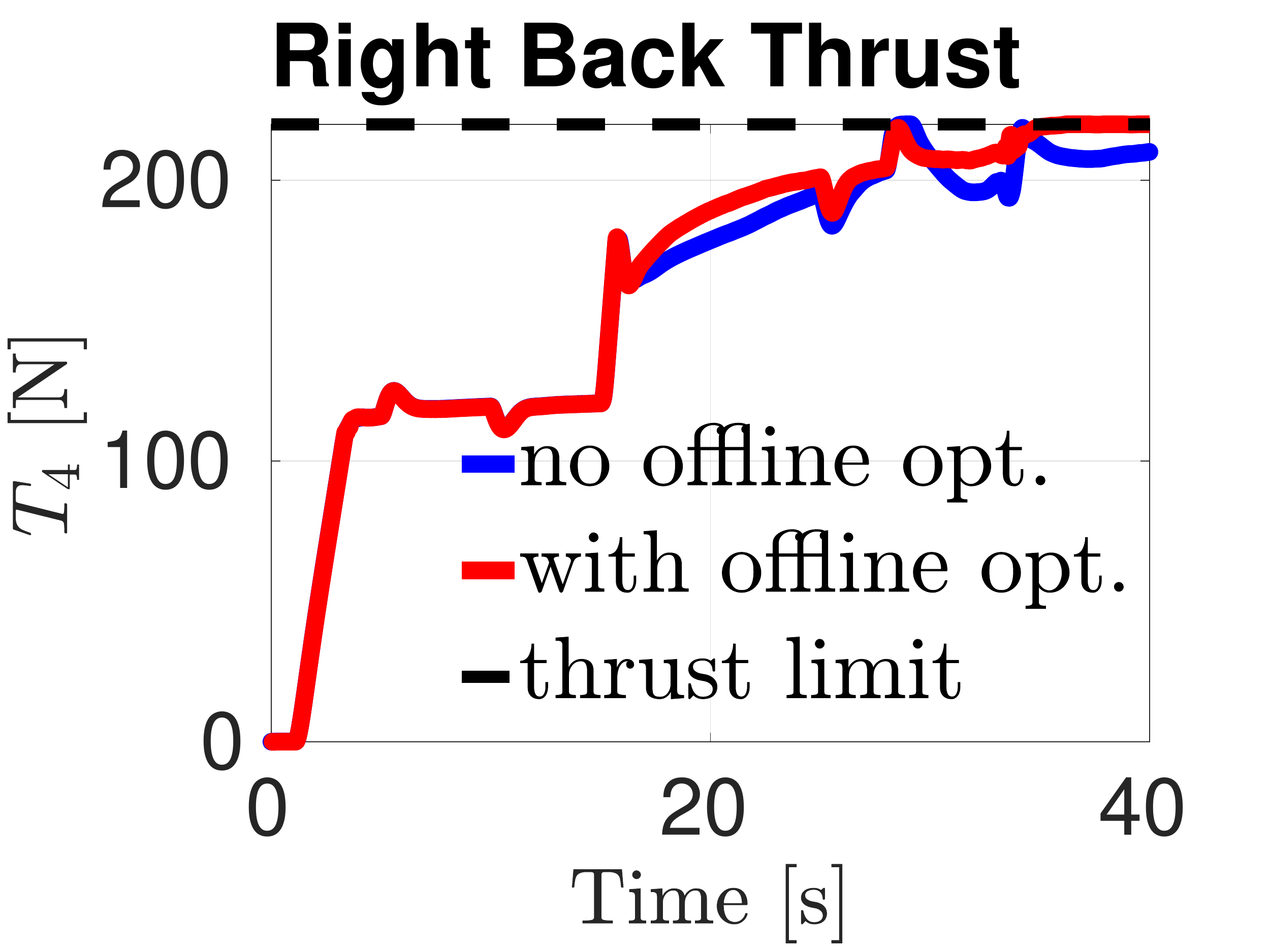}
   \end{minipage}
   \begin{minipage}[c]{8.5cm}
     \centering
     \includegraphics[height=3.8cm, width=\columnwidth]{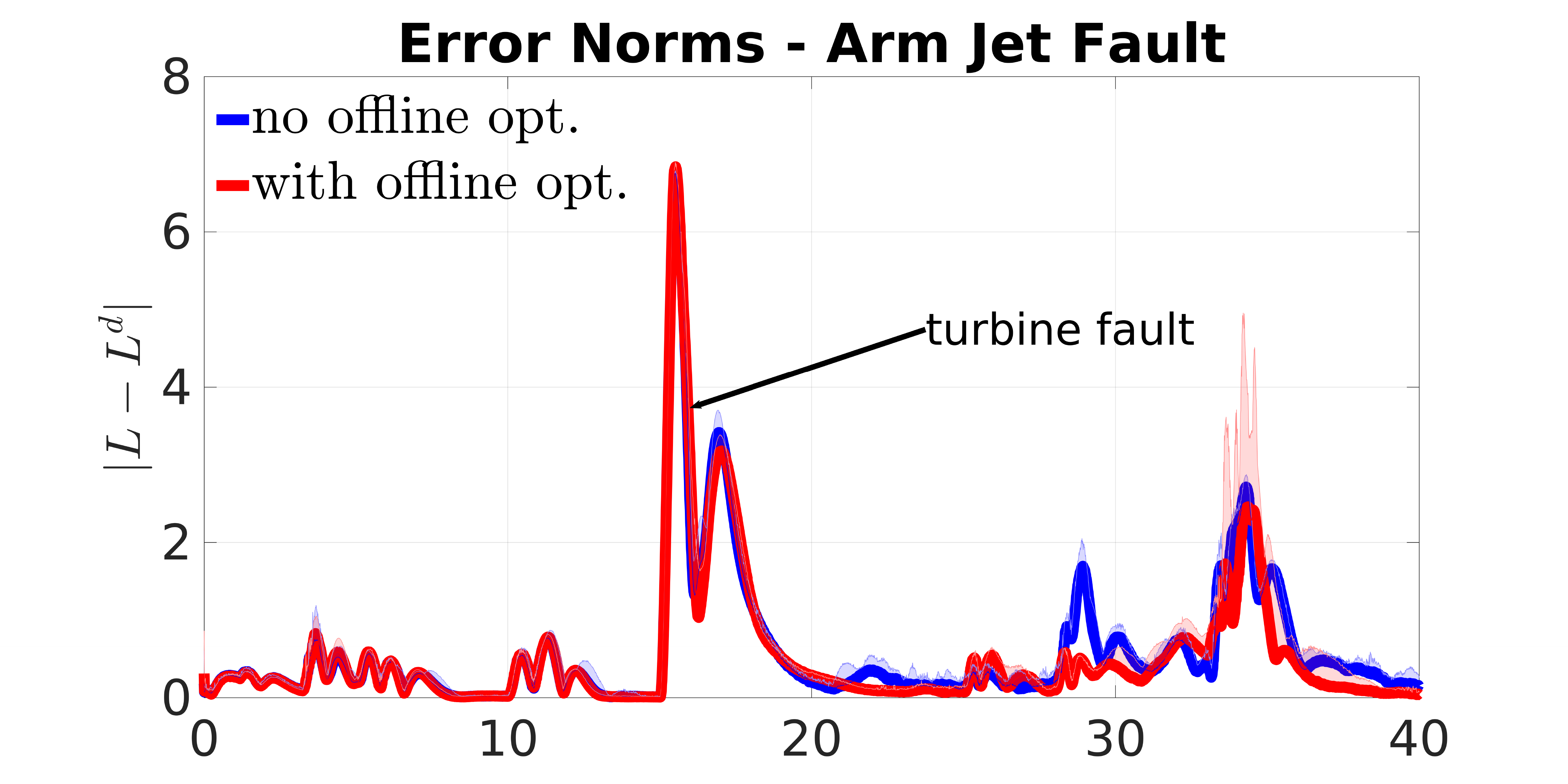}
     \vspace{-0.45cm}
   \end{minipage}
   \begin{minipage}[c]{8.5cm}
     \centering
     \includegraphics[height=3.8cm, width=\columnwidth]{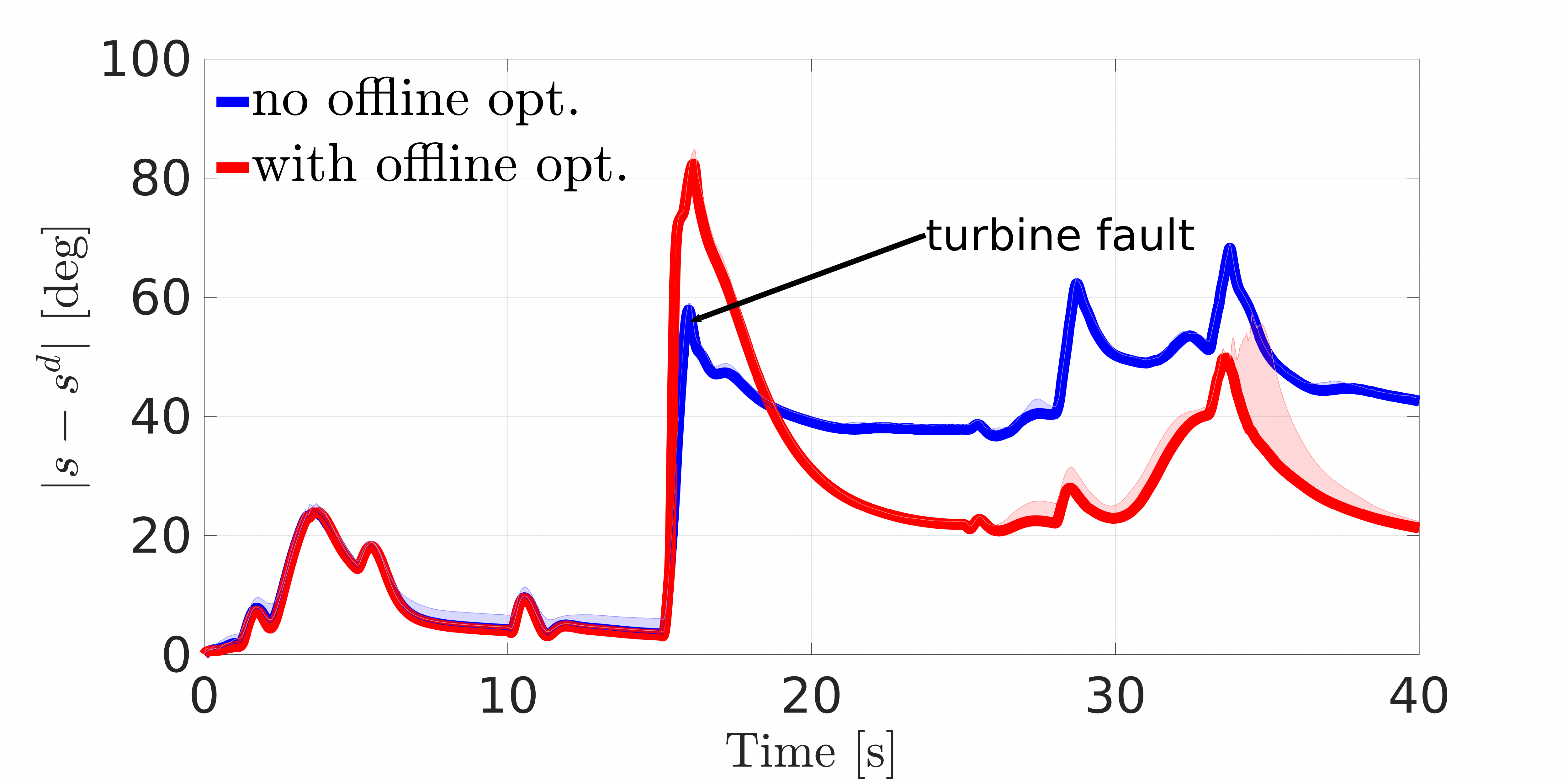}
     \vspace{-0.55cm}
     \caption{Jet thrusts and error norms of QP tasks with arm turbine fault. From top to bottom: thrust intensities during the simulated trajectory; momentum error norm; joint positions error norm.}
     \label{fig:qpTasksNormArm}
   \end{minipage}
   \vspace{-0.5cm}
\end{figure} 

\begin{figure}[t]
   \begin{minipage}[c]{4.25cm}
     \centering
     \includegraphics[height=2.5cm, width=0.9\columnwidth]{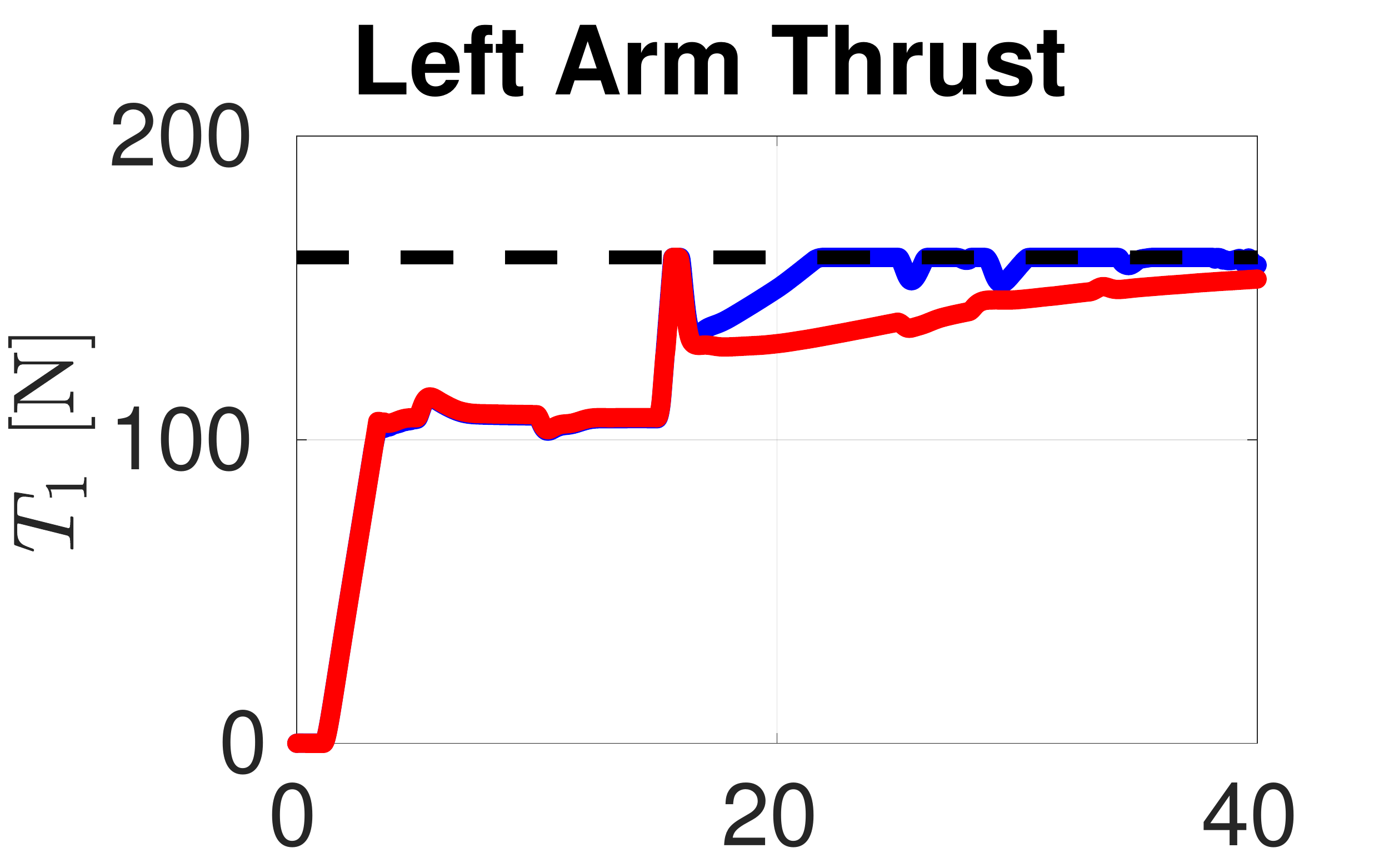}
     \hspace{-0.5cm}
   \end{minipage}
   \begin{minipage}[c]{4.25cm}
     \includegraphics[height=2.5cm, width=0.9\columnwidth]{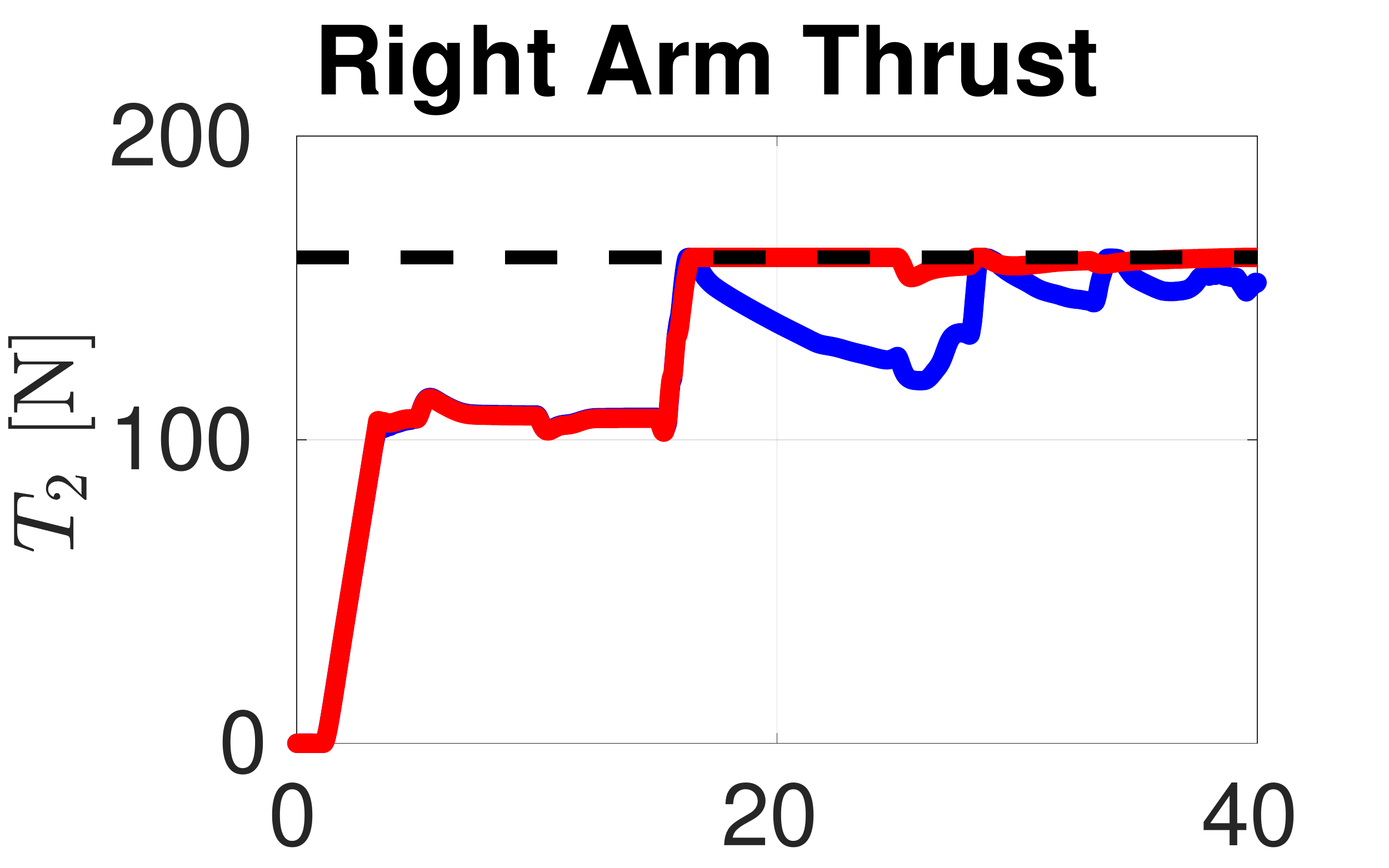}
   \end{minipage}
   \begin{minipage}[c]{4.25cm}
     \centering
     \includegraphics[height=2.95cm, width=0.9\columnwidth]{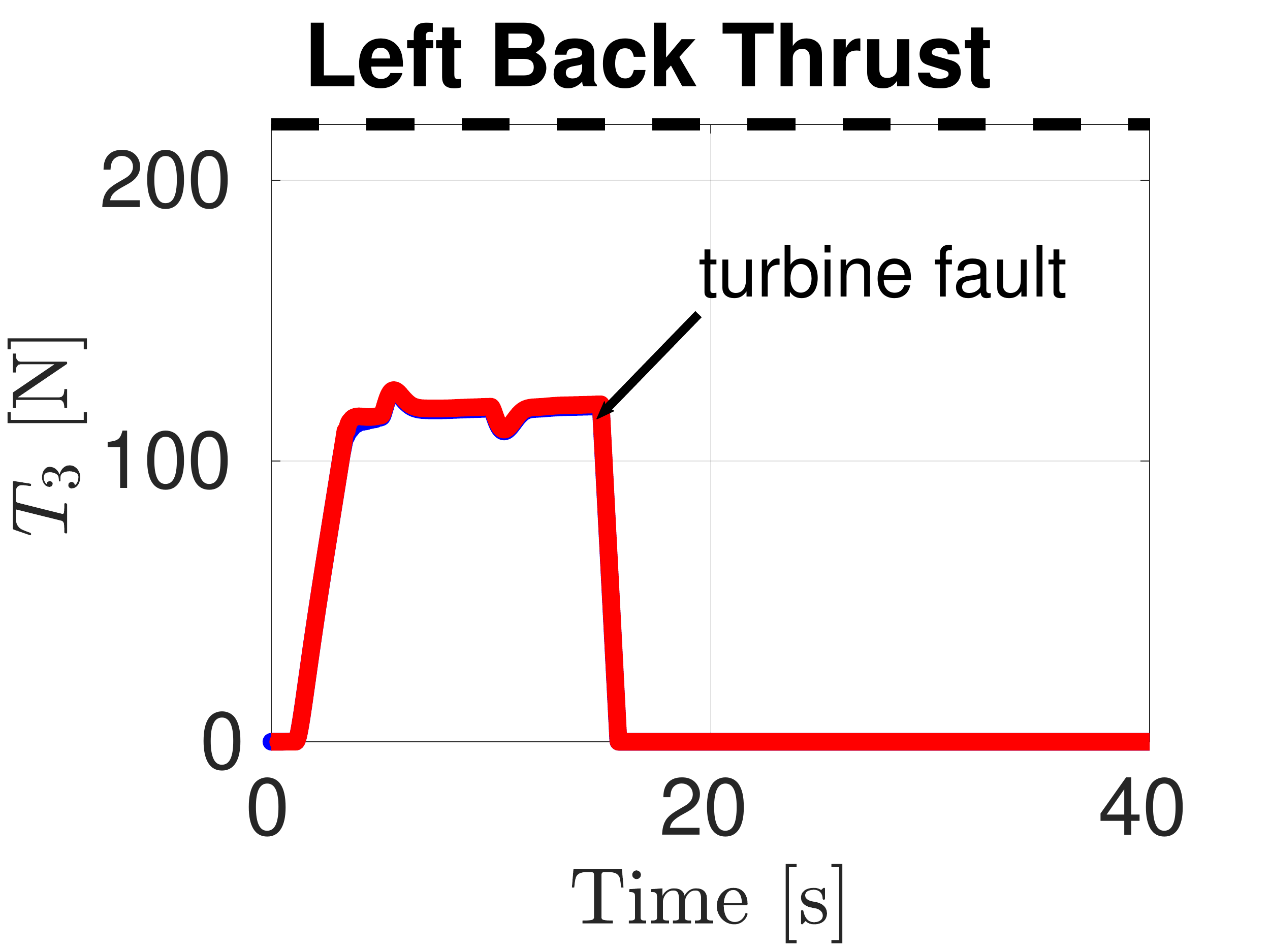}
     \hspace{-0.5cm}
   \end{minipage}
   \begin{minipage}[c]{4.25cm}
     \includegraphics[height=2.95cm, width=0.9\columnwidth]{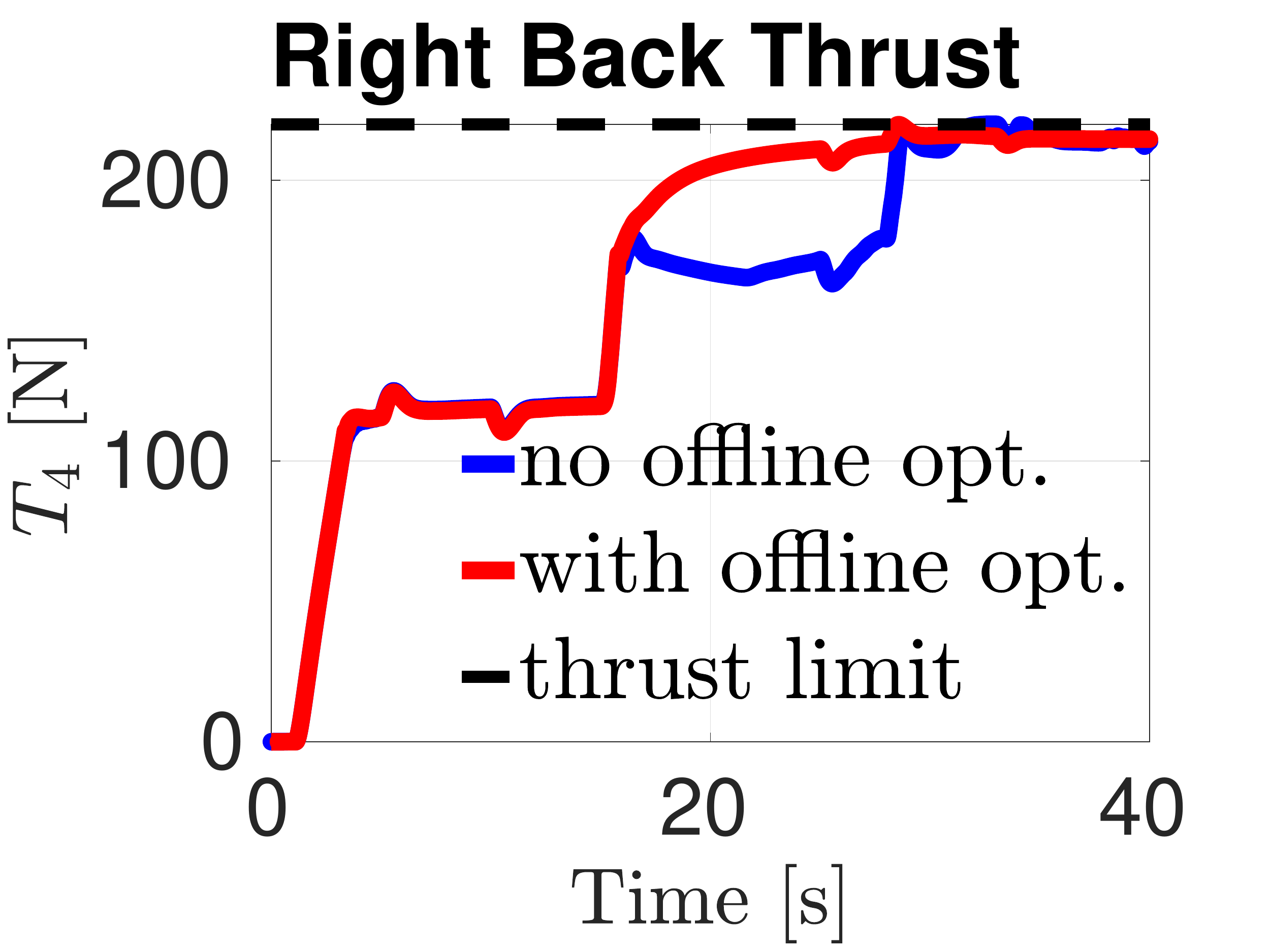}
   \end{minipage}
   \begin{minipage}[c]{8.5cm}
     \centering
     \includegraphics[height=3.8cm, width=\columnwidth]{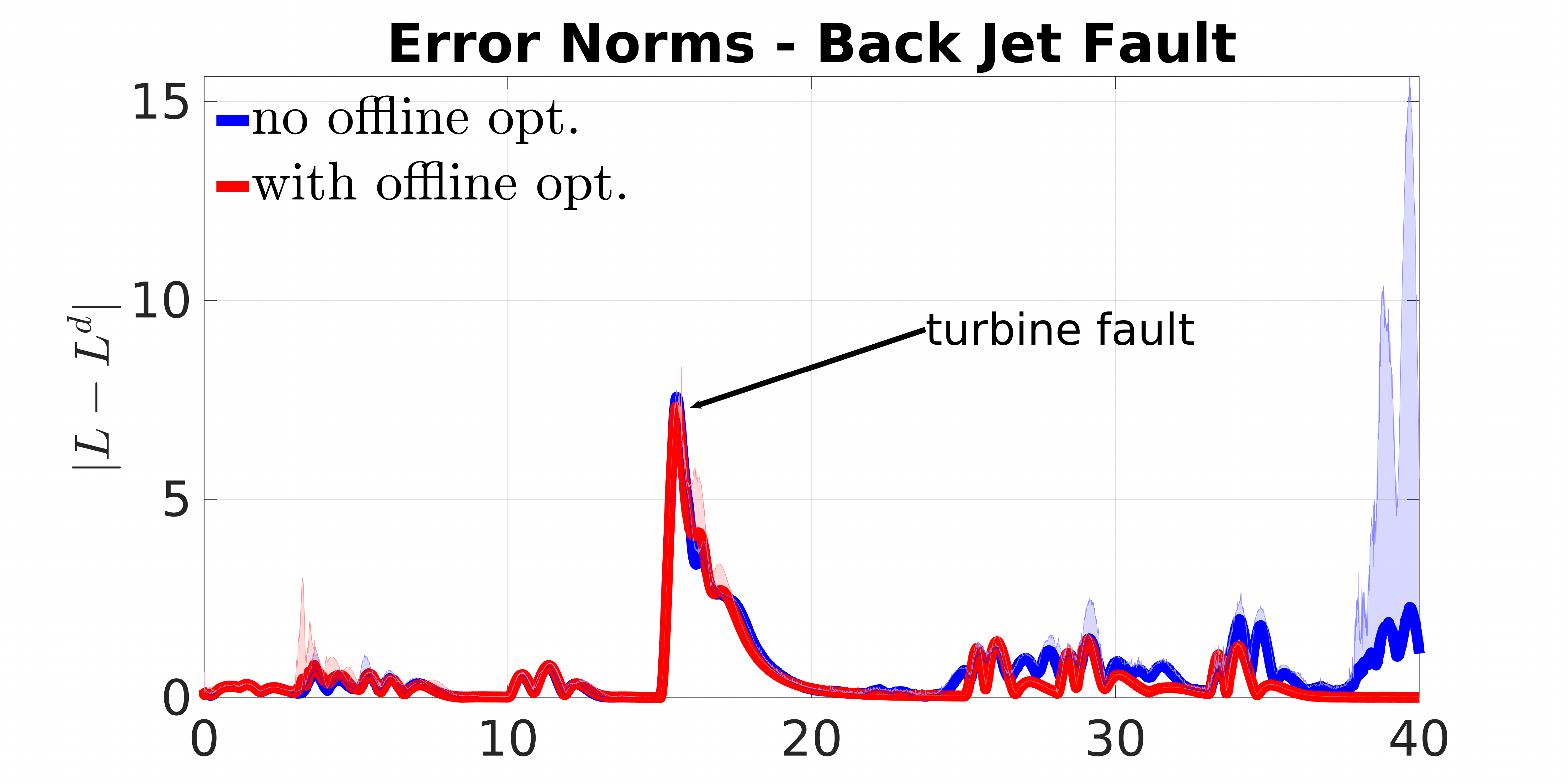}
     \vspace{-0.45cm}
   \end{minipage}
   \begin{minipage}[c]{8.5cm}
     \centering
     \includegraphics[height=3.8cm, width=\columnwidth]{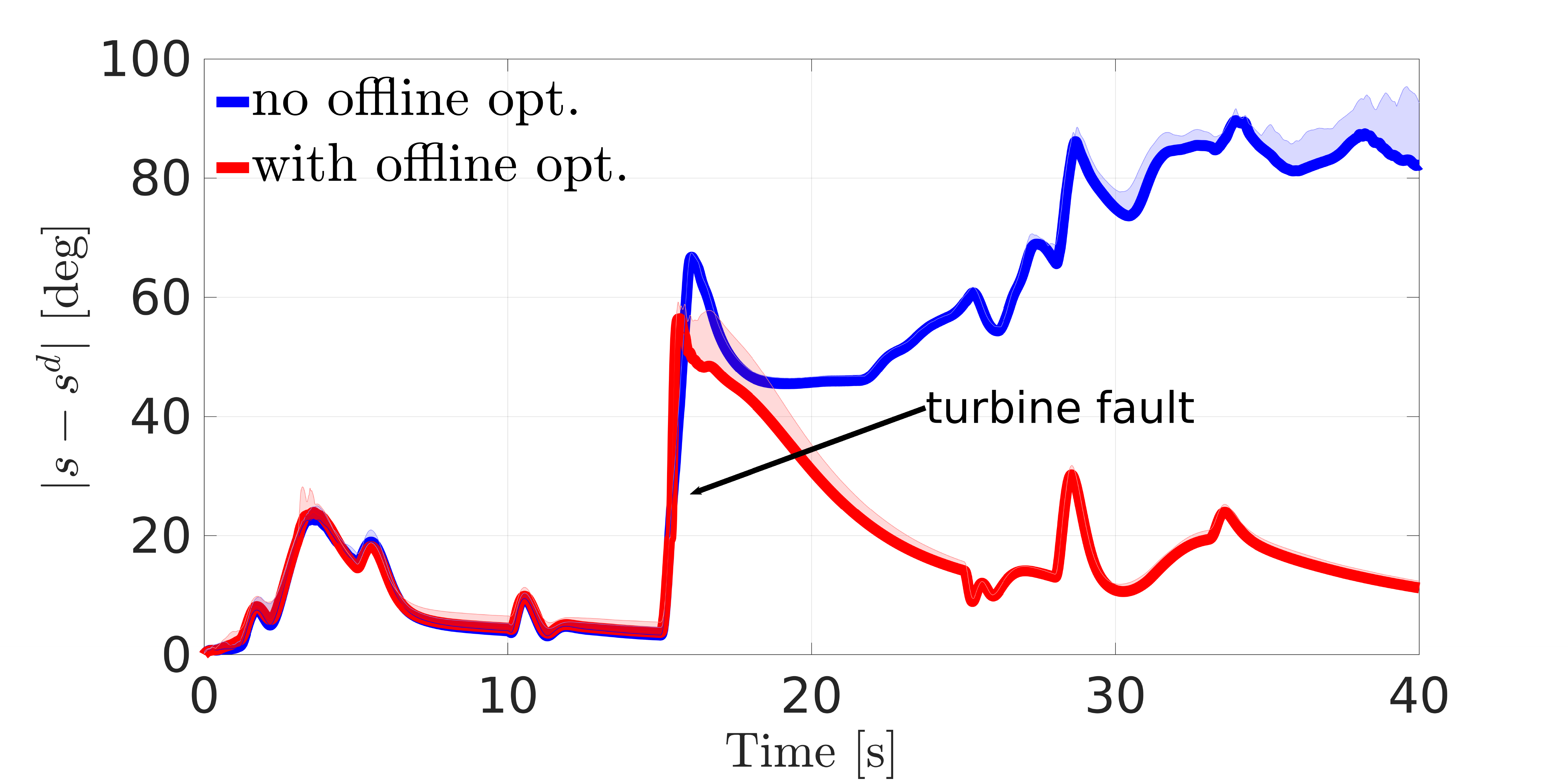}
     \vspace{-0.55cm}
     \caption{Jet thrusts and error norms of QP tasks with back turbine fault. From top to bottom: thrust intensities during the simulated trajectory; momentum error norm; joint positions error norm.}
     \label{fig:qpTasksNormBack}
   \end{minipage}
   \vspace{-0.5cm}
\end{figure}

The top part of Figs. \ref{fig:qpTasksNormArm} and \ref{fig:qpTasksNormBack} depict the thrust intensities in case of arm and back turbine failure, respectively. On the bottom part we show the error norms of linear and angular momentum (combined together) and joint positions. The thick blue and red lines represent the average over $10$ experiments, while the shadowed areas indicate the maximum error norm reached at each time instant among all simulations. Results in blue are obtained without using the reference generator of Sec. \ref{sec:ref_gen}, while results in red also include updated attitude and joint positions references. For the arm turbine failure case Fig. \ref{fig:qpTasksNormArm}, the fault controller alone (blue lines) is already capable of driving the momentum error norm to zero, and keeps bounded the norm of the lower priority task. The addition of the reference generator slightly improves the average controller performances in tracking all tasks. For what concerns instead the back turbine failure (Fig. \ref{fig:qpTasksNormBack}), the controller without reference generator struggles to converge the momentum error to zero, and the joints error keeps increasing during the trajectory tracking task. This behaviour confirms the observations done in Sec. \ref{subsec:controller_limits} about the need of providing the controller with far-from-singularities and optimized references after the failure occurs. In fact, when the reference generator is also included (red lines), performances are considerably improved. In all simulations, the robot was able to resist the failure and complete the flight. 

The code used throughout the paper is stored in a public repository: \url{https://github.com/ami-iit/paper_nava_2023_icra_fault-control-ironcub}. The achieved results can be reproduced once all the necessary dependencies have been installed.

\section{CONCLUSIONS}
\label{sec:conclusions}

We proposed a fault detection and control strategy for a multi-body flying robot equipped with jet engines. Our framework includes: a RPM-based fault detection system, modifications of the existing momentum-based flight controller to recover stability after a jet failure, and an offline reference generator to further improve robustness and performances of the fault control. Simulations proved the effectiveness of our approach and highlighted the importance of the reference generator to maintain a stable flight after the fault occurs.

At the current stage, the robot cannot land properly after a failure because its feet do not remain parallel to the ground (although it could be driven close to the landing site and forced to land, avoiding major damages). Furthermore, some parameters related to jet engines (e.g., jet exhausts profile, jet dynamics) still need to be estimated appropriately and integrated in the control design. Future work will investigate in the above-mentioned directions. We also aim to test and improve our algorithm on the real iRonCub, for which flight experiments are currently on going.

\addtolength{\textheight}{0cm}     

\bibliographystyle{IEEEtran}
\bibliography{IEEEabrv,Biblio}

\begin{thebibliography}{10}
\providecommand{\url}[1]{#1}
\csname url@rmstyle\endcsname
\providecommand{\newblock}{\relax}
\providecommand{\bibinfo}[2]{#2}
\providecommand\BIBentrySTDinterwordspacing{\spaceskip=0pt\relax}
\providecommand\BIBentryALTinterwordstretchfactor{4}
\providecommand\BIBentryALTinterwordspacing{\spaceskip=\fontdimen2\font plus
\BIBentryALTinterwordstretchfactor\fontdimen3\font minus
  \fontdimen4\font\relax}
\providecommand\BIBforeignlanguage[2]{{%
\expandafter\ifx\csname l@#1\endcsname\relax
\typeout{** WARNING: IEEEtran.bst: No hyphenation pattern has been}%
\typeout{** loaded for the language `#1'. Using the pattern for}%
\typeout{** the default language instead.}%
\else
\language=\csname l@#1\endcsname
\fi
#2}}

\bibitem{Yin2016}
S.~Yin, B.~Xiao, S.~X. Ding, and D.~Zhou, ``A review on recent development of
  spacecraft attitude fault tolerant control system,'' \emph{IEEE Transactions
  on Industrial Electronics}, vol.~63, no.~5, pp. 3311--3320, 2016.

\bibitem{Zhang2003}
Y.~Zhang and J.~Jiang, ``Bibliographical review on reconfigurable
  fault-tolerant control systems,'' \emph{IFAC Proceedings Volumes}, vol.~36,
  pp. 257--268, 06 2003.

\bibitem{fourlas2021}
\BIBentryALTinterwordspacing
G.~K. Fourlas and G.~C. Karras, ``A survey on fault diagnosis and
  fault-tolerant control methods for unmanned aerial vehicles,''
  \emph{Machines}, vol.~9, no.~9, 2021. [Online]. Available:
  \url{https://www.mdpi.com/2075-1702/9/9/197}
\BIBentrySTDinterwordspacing

\bibitem{jun2010}
W.~Jun, Y.~Xiong-Dong, and T.~Yu-Yang, ``Fault-tolerant control design of
  quadrotor uav based on cpso,'' in \emph{2018 IEEE 4th International
  Conference on Control Science and Systems Engineering (ICCSSE)}, 2018, pp.
  279--283.

\bibitem{gong2010}
W.~Gong, J.~Zhang, B.~Li, and Y.~Yang, ``Integral-type sliding mode based
  fault-tolerant attitude stabilization of a quad-rotor uav,'' in \emph{2018
  International Symposium in Sensing and Instrumentation in IoT Era (ISSI)},
  2018, pp. 1--6.

\bibitem{mallavalli2019}
S.~Mallavalli and A.~Fekih, ``A fault tolerant control design for actuator
  fault mitigation in quadrotor uavs,'' in \emph{2019 American Control
  Conference (ACC)}, 2019, pp. 5111--5116.

\bibitem{yu2021}
Z.~Yu, Y.~Zhang, B.~Jiang, C.-Y. Su, J.~Fu, Y.~Jin, and T.~Chai,
  ``Fractional-order adaptive fault-tolerant synchronization tracking control
  of networked fixed-wing uavs against actuator-sensor faults via intelligent
  learning mechanism,'' \emph{IEEE Transactions on Neural Networks and Learning
  Systems}, vol.~32, no.~12, pp. 5539--5553, 2021.

\bibitem{Sadeghzadeh2013}
I.~Sadeghzadeh and Y.~Zhang, ``Actuator fault-tolerant control based on
  gain-scheduled pid with application to fixed-wing unmanned aerial vehicle,''
  in \emph{2013 Conference on Control and Fault-Tolerant Systems (SysTol)},
  2013, pp. 342--346.

\bibitem{Cheng2018}
P.~Cheng, Z.~Gao, M.~Qian, and J.~Lin, ``Active fault tolerant control design
  for uav using nonsingular fast terminal sliding mode approach,'' in
  \emph{2018 Chinese Control And Decision Conference (CCDC)}, 2018, pp.
  292--297.

\bibitem{Yu2015}
B.~Yu, Y.~Zhang, and Y.~Qu, ``Mpc-based ftc with fdd against actuator faults of
  uavs,'' in \emph{2015 15th International Conference on Control, Automation
  and Systems (ICCAS)}, 2015, pp. 225--230.

\bibitem{zhong2018}
Y.~Zhong, Y.~Zhang, and W.~Zhang, ``Active fault-tolerant tracking control of a
  quadrotor uav,'' in \emph{2018 International Conference on
  Sensing,Diagnostics, Prognostics, and Control (SDPC)}, 2018, pp. 497--502.

\bibitem{Nguyen2018}
D.-T. Nguyen, D.~Saussié, and L.~Saydy, ``Fault-tolerant control of a
  hexacopter uav based on self-scheduled control allocation,'' in \emph{2018
  International Conference on Unmanned Aircraft Systems (ICUAS)}, 2018, pp.
  385--393.

\bibitem{kim2021}
K.~Kim, P.~Spieler, E.~Lupu, A.~Ramezani, and S.~Chung, ``A bipedal walking
  robot that can fly, slackline, and skateboard,'' \emph{Science Robotics},
  vol.~6, no.~59, 2021.

\bibitem{huang2017}
Z.~{Huang}, B.~{Liu}, J.~{Wei}, Q.~{Lin}, J.~{Ota}, and Y.~{Zhang}, ``Jet-hr1:
  Two-dimensional bipedal robot step over large obstacle based on a ducted-fan
  propulsion system,'' in \emph{2017 IEEE-RAS 17th International Conference on
  Humanoid Robotics (Humanoids)}, 2017.

\bibitem{ruggiero2018}
F.~{Ruggiero}, V.~{Lippiello}, and A.~{Ollero}, ``Aerial manipulation: A
  literature review\looseness=-1,'' \emph{IEEE Robotics and Automation
  Letters\looseness=-1}, vol.~3, no.~3, pp. 1957--1964, July 2018.

\bibitem{kalantari2013}
A.~Kalantari and M.~Spenko, ``Design and experimental validation of hytaq, a
  hybrid terrestrial and aerial quadrotor,'' in \emph{2013 IEEE International
  Conference on Robotics and Automation}, May 2013, pp. 4445--4450.

\bibitem{pitonyak2017}
M.~Pitonyak and F.~Sahin, ``A novel hexapod robot design with flight
  capability,'' in \emph{2017 12th System of Systems Engineering Conference
  (SoSE)}, June 2017, pp. 1--6.

\bibitem{pose2022}
C.~Pose, J.~Giribet, and I.~Mas, ``Adaptive center-of-mass relocation for
  aerial manipulator fault tolerance,'' \emph{IEEE Robotics and Automation
  Letters}, vol.~7, no.~2, pp. 5583--5590, 2022.

\bibitem{garimella2021}
G.~Garimella, M.~Sheckells, S.~Kim, G.~Baraban, and M.~Kobilarov, ``Improving
  the reliability of pick-and-place with aerial vehicles through fault-tolerant
  software and a custom magnetic end-effector,'' \emph{IEEE Robotics and
  Automation Letters}, vol.~6, no.~4, pp. 7501--7508, 2021.

\bibitem{pucci2017fly}
D.~Pucci, S.~Traversaro, and F.~Nori, ``Momentum control of an underactuated
  flying humanoid robot,'' \emph{IEEE Robotics and Automation Letters}, vol.~3,
  no.~1, pp. 195--202, Jan 2018.

\bibitem{mohamed2020}
H.~A.~O. Mohamed, G.~Nava, G.~L’Erario, S.~Traversaro, F.~Bergonti,
  L.~Fiorio, P.~R. Vanteddu, F.~Braghin, and D.~Pucci, ``Momentum-based
  extended kalman filter for thrust estimation on flying multibody robots,''
  \emph{IEEE Robotics and Automation Letters}, vol.~7, no.~1, pp. 526--533,
  2022.

\bibitem{Bartolozzi2017}
\BIBentryALTinterwordspacing
L.~Natale, C.~Bartolozzi, D.~Pucci, A.~Wykowska, and G.~Metta, ``icub: The
  not-yet-finished story of building a robot child,'' \emph{Science Robotics},
  vol.~2, no.~13, 2017. [Online]. Available:
  \url{https://robotics.sciencemag.org/content/2/13/eaaq1026}
\BIBentrySTDinterwordspacing

\bibitem{nava2018}
G.~Nava, L.~Fiorio, S.~Traversaro, and D.~Pucci, ``Position and attitude
  control of an underactuated flying humanoid robot,'' in \emph{2018 IEEE-RAS
  18th International Conference on Humanoid Robots (Humanoids)}, 2018, pp.
  1--9.

\bibitem{Featherstone2007}
R.~Featherstone, \emph{Rigid Body Dynamics Algorithms}.\hskip 1em plus 0.5em
  minus 0.4em\relax Secaucus, NJ, USA: Springer-Verlag New York, Inc., 2007.

\bibitem{Marsden2010}
J.~E. Marsden and T.~S. Ratiu, \emph{Introduction to Mechanics and Symmetry: A
  Basic Exposition of Classical Mechanical Systems}.\hskip 1em plus 0.5em minus
  0.4em\relax Springer Publishing Company, Incorporated, 2010.

\bibitem{Acosta05}
J.~A. Acosta and M.~Lopez-Martinez, ``{Constructive feedback linearization of
  underactuated mechanical systems with 2-DOF},'' \emph{Decision and Control,
  2005 and 2005 European Control Conference. CDC-ECC '05. 44th IEEE Conference
  on}, 2005.

\bibitem{Orin2008}
D.~E. Orin and A.~Goswami, ``Centroidal momentum matrix of a humanoid robot:
  Structure and properties,'' in \emph{2008 IEEE/RSJ International Conference
  on Intelligent Robots and Systems}, 2008, pp. 653--659.

\bibitem{nava2020_thesis}
\BIBentryALTinterwordspacing
G.~Nava, \emph{Instantaneous Momentum-Based Control of Floating Base
  Systems}.\hskip 1em plus 0.5em minus 0.4em\relax Ph.D. Thesis, 2020.
  [Online]. Available: \url{https://iris.unige.it/handle/11567/1004907#}
\BIBentrySTDinterwordspacing

\bibitem{Rapetti2020}
\BIBentryALTinterwordspacing
L.~Rapetti, Y.~Tirupachuri, K.~Darvish, S.~Dafarra, G.~Nava, C.~Latella, and
  D.~Pucci, ``Model-based real-time motion tracking using dynamical inverse
  kinematics,'' \emph{Algorithms}, vol.~13, no.~10, 2020. [Online]. Available:
  \url{https://www.mdpi.com/1999-4893/13/10/266}
\BIBentrySTDinterwordspacing

\bibitem{Momin2022}
A.~J.~A. Momin, G.~Nava, G.~L'Erario, H.~A.~O. Mohamed, F.~Bergonti, P.~R.
  Vanteddu, F.~Braghin, and D.~Pucci, ``Nonlinear model identification and
  observer design for thrust estimation of small-scale turbojet engines,'' in
  \emph{2022 International Conference on Robotics and Automation (ICRA)}, 2022,
  pp. 5879--5885.

\bibitem{Koenig04}
N.~Koenig and A.~Howard, ``Design and use paradigms for gazebo, an open-source
  multi-robot simulator,'' \emph{Intelligent Robots and Systems, 2004. (IROS
  2004). Proceedings. 2004 IEEE/RSJ International Conference on}, pp. 2149 --
  2154, 2004.

\bibitem{Metta2006}
\BIBentryALTinterwordspacing
G.~Metta, P.~Fitzpatrick, and L.~Natale, ``Yarp: Yet another robot platform,''
  \emph{International Journal of Advanced Robotic Systems}, vol.~3, no.~1,
  p.~8, 2006. [Online]. Available: \url{https://doi.org/10.5772/5761}
\BIBentrySTDinterwordspacing

\end{thebibliography}

\end{document}